\documentclass[lettersize,journal]{IEEEtran}
\usepackage{times}  
\usepackage{helvet}  
\usepackage{courier}  
\usepackage[hyphens]{url}  
\usepackage{graphicx} 
\usepackage{natbib}  
\usepackage{caption} 
\usepackage{CJKutf8}
\usepackage{algorithmic}
\usepackage{booktabs}
\usepackage{multirow}
\usepackage{comment}
\usepackage{xcolor}
\usepackage[table]{xcolor}
\usepackage{subcaption}
\usepackage{hyperref}
\usepackage[accsupp]{axessibility}  
\usepackage{newfloat}
\usepackage{listings}
\usepackage{graphicx}
\usepackage{amsmath}
\usepackage{amssymb}
\usepackage[linesnumbered,ruled,vlined]{algorithm2e}
\usepackage{algorithm2e}
\usepackage{CJKutf8}
\usepackage{makecell}
\usepackage{wrapfig}

\usepackage{xcolor}
\usepackage{booktabs} 
\usepackage{multirow}

\usepackage{array}
\setcitestyle{numbers,square,comma}

\title{Pixel-Space Diffusion Transformers}
\author{
    Renye Yan, Jikang Cheng, You Wu, Ling Liang, Wei Peng$^\dagger$, Athanasios V. Vasilakos, Qingyu Zhao, Yu Zhang, Yimao Cai, Kilian M. Pohl, Guoying Zhao \textit{Fellow, IEEE }
    \thanks{
    Renye Yan, Jikang Cheng, Ling Liang, and Yimao Cai are with Peking University, China. (e-mail: victory@stu.pku.edu.cn, jikangcheng99@gmail.com. 

    You Wu is with Nanjing University, China.

    Wei Peng, Yu Zhang, and Kilian M. Pohl are with Stanford University, USA.

    Qingyu Zhao is with Cornell University, USA.

    Athanasios V. Vasilakos is With the Center for AI Research,University of Agder(UiA), Norway.

    Guoying Zhao is with the University of Oulu, Finland.
    
    $^\dagger$Corresponding author.
    }
    
}

\usepackage{bibentry}

\usepackage{array}
\usepackage{tabularx}
\usepackage[table]{xcolor}
\usepackage{caption}

\definecolor{TitleBlue}{HTML}{172C67}
\definecolor{LineBlue}{HTML}{AAB8E8}

\definecolor{BlueText}{HTML}{1E55B3}
\definecolor{GreenText}{HTML}{287B32}
\definecolor{OrangeText}{HTML}{EE6B19}
\definecolor{PurpleText}{HTML}{7040A7}
\definecolor{CyanText}{HTML}{13879A}
\definecolor{RedText}{HTML}{D52D35}
\definecolor{DeepBlueText}{HTML}{164D9B}

\definecolor{BlueBG}{HTML}{F5F8FF}
\definecolor{GreenBG}{HTML}{F5FAF4}
\definecolor{OrangeBG}{HTML}{FFF9F3}
\definecolor{PurpleBG}{HTML}{FAF7FD}
\definecolor{CyanBG}{HTML}{F3FAFC}
\definecolor{RedBG}{HTML}{FFF7F7}

\newcommand{\CenterCell}[1]{%
\begin{minipage}[t][2.55cm][t]{\linewidth}
\centering
\vspace{0.05cm}
#1
\end{minipage}%
}

\newcommand{\LeftCell}[1]{%
\begin{minipage}[t][2.55cm][t]{\linewidth}
\raggedright
\vspace{0.05cm}
#1
\end{minipage}%
}

\newcommand{\strength}[1]{%
\textcolor{BlueText}{\textbf{Strength: }}#1%
}

\newcommand{\limitation}[1]{%
\textcolor{GreenText}{\textbf{Limitation: }}#1%
}

\definecolor{ChallengeBlue}{HTML}{183A85}
\definecolor{EssenceGreen}{HTML}{69B34C}
\definecolor{SolutionPurple}{HTML}{7052B8}

\definecolor{ChallengeBG}{HTML}{F6F8FD}
\definecolor{EssenceBG}{HTML}{F7FBF4}
\definecolor{SolutionBG}{HTML}{FAF8FD}
\definecolor{GoalBG}{HTML}{F8FAFE}

\definecolor{FormulaGray}{HTML}{303030}
\definecolor{SolutionText}{HTML}{6943B5}

\newcommand{\ChallengeCell}[1]{%
\begin{minipage}[t][3.10cm][t]{\linewidth}
\centering
\vspace{0.14cm}
\textcolor{ChallengeBlue}{\bfseries #1}
\end{minipage}%
}

\newcommand{\EssenceCell}[1]{%
\begin{minipage}[t][3.10cm][t]{\linewidth}
\raggedright
\vspace{0.10cm}
#1
\end{minipage}%
}

\newcommand{\SolutionCell}[1]{%
\begin{minipage}[t][3.10cm][t]{\linewidth}
\raggedright
\vspace{0.10cm}
#1
\end{minipage}%
}

\newcommand{\solutionitem}[2]{%
\textcolor{SolutionText}{\bfseries #1：}#2%
}

\newcommand{\tablebullet}{%
\raisebox{0.15ex}{\scriptsize$\bullet$}\hspace{0.45em}%
}

\begin{document}
\begin{CJK*}{UTF8}{gbsn}
\maketitle

\begin{abstract}
Latent diffusion models enable efficient image synthesis by performing denoising in a VAE-compressed latent space. However, this VAE paradigm becomes increasingly restrictive as modern visual generation demands finer textures, anatomically plausible subjects, and coherent object relationships. Fixed visual tokenizers may discard fine-grained information that cannot be recovered by the downstream diffusion model. Moreover, separating representation learning from diffusion training creates a mismatch between reconstruction and generation objectives, limiting end-to-end optimization and often requiring additional effort to align the latent space with generative objectives. These limitations have renewed interest in Pixel-Space Diffusion, which directly models raw pixels, allowing the diffusion model to operate on uncompressed visual data and jointly optimize representation learning and generation. This formulation is better aligned with the quality requirements of modern high-fidelity visual generation.  Pixel-space modeling further offers a promising foundation for unified multimodal models: raw pixels, text, and task conditions can be represented in a shared token space and jointly modeled by a single Transformer, narrowing the gap between visual understanding and generation. However, it also introduces challenges in high-dimensional modeling, including noise scheduling, loss weighting, token efficiency, and scalable architecture design. This paper reviews Pixel-Space Diffusion Transformers from the perspectives of model architectures, continuous generative mechanisms, and unified multimodal modeling. We summarize key methods, identify core challenges, and discuss future directions toward high-fidelity, end-to-end vision foundation models that integrate generation and understanding.
\end{abstract}

\section{Introduction}
With the rapid development of computer vision~\cite{cheng2026sanity,xu2026edge,kage2026review,ballesteros2026intelligent,ji2026computer,xu2026edge,kage2026review,guo2022attention,wang2021generative}, generative models~\citep{heitmann2026picture,dufour2025around,yan2025entropy,bond2021deep,zhao2016energy,dinh2014nice,van2016wavenet,zhu2017unpaired} have emerged as a foundational technical framework that bridging data understanding, content creation, and interactive applications. As visual data scales and generative artificial intelligence advances, models are expected not only to synthesize highly realistic images, but also to seamlessly handle complex conditional control, preserve fine-grained structural features, and facilitate fluid cross-modal interactions.


Among various generative modeling paradigms, Diffusion Models~\cite{croitoru2023diffusion,zou2026mixture,wang2026remasking,bonnaire2026diffusion,karras2024guiding,tevet2022human,fuest2026diffusion,karnewar2023holodiffusion,li2026videococo} are the dominant technical paradigm for image generation~\citep{bonnaire2026diffusion,kim2026klass,zampini2026training,yan2026less}. 
By progressively learning to reverse a noising process to recover the underlying data distribution, these models synthesize high-quality visual samples. However, performing denoising directly in the image pixel space introduces a severe computational bottleneck. To bypass this hurdle, Latent Diffusion Models (LDMs)~\citep{rombach2022high,podell2024sdxl,blattmann2023align,blattmann2023stable,kwon2022diffusion,corneanu2024latentpaint,dar2026unconditional} introduced a two-stage ``compress-then-diffuse" strategy: by leveraging a pre-trained Variational Autoencoder (VAE)~\citep{sonderby2016ladder,ramchandran2021longitudinal,kusner2017grammar,yan2024exploration,khattar2019mvae,vafaii2024poisson,abdulganiyu2025xidintfl,zhang2026remaining} to map raw images into a compact, low-dimensional latent space, followed by second-stage denoising-based generation. Through this staged design, LDMs drastically reduce the computational footprint of high-resolution synthesis.



Despite LDMs' undeniable efficiency, the aforementioned two-stage paradigm imposes a rigid performance ceiling dictated by its structural design~\cite{crowson2024scalable,atzmon2024edify,rombach2022high}. 
Specifically, lossy spatial compression within the VAE routinely discards high-frequency textures, sharp boundaries, and minute local structures. This creates an irreversible information bottleneck and as each of the LDMs two stages have distinct optimizing objectives, which causes a fundamental representation mismatch between the autoencoder's reconstruction criteria and the diffusion model's generative objectives~\cite{ma2026pixelgen,wang2025pixnerd,yu2026pixeldit}.

Driven by these limitations, novel Pixel-Space Diffusion~\cite{baade2026latent,bradbury2026your,mukhopadhyay2026scale,nguyen2026pixel,hoogeboom2025simpler,xu2026pixel,elata2025novel,zhang2024pixel,jain2023vectorfusion} has recently experienced an algorithmic renaissance. Far from a nostalgic regression to early convolutional pixel-level models, this movement represents a sophisticated re-examination of end-to-end visual modeling. Powered by scalable Transformer architectures, continuous generative formulations, and massive distributed compute, modern pixel-space research aims to eliminate the structural bottlenecks of external VAEs without entirely dismissing their historical insights on efficiency~\cite{zhang2024pixel}.

\begin{figure*}
    \centering
    \includegraphics[width=1.0\linewidth]{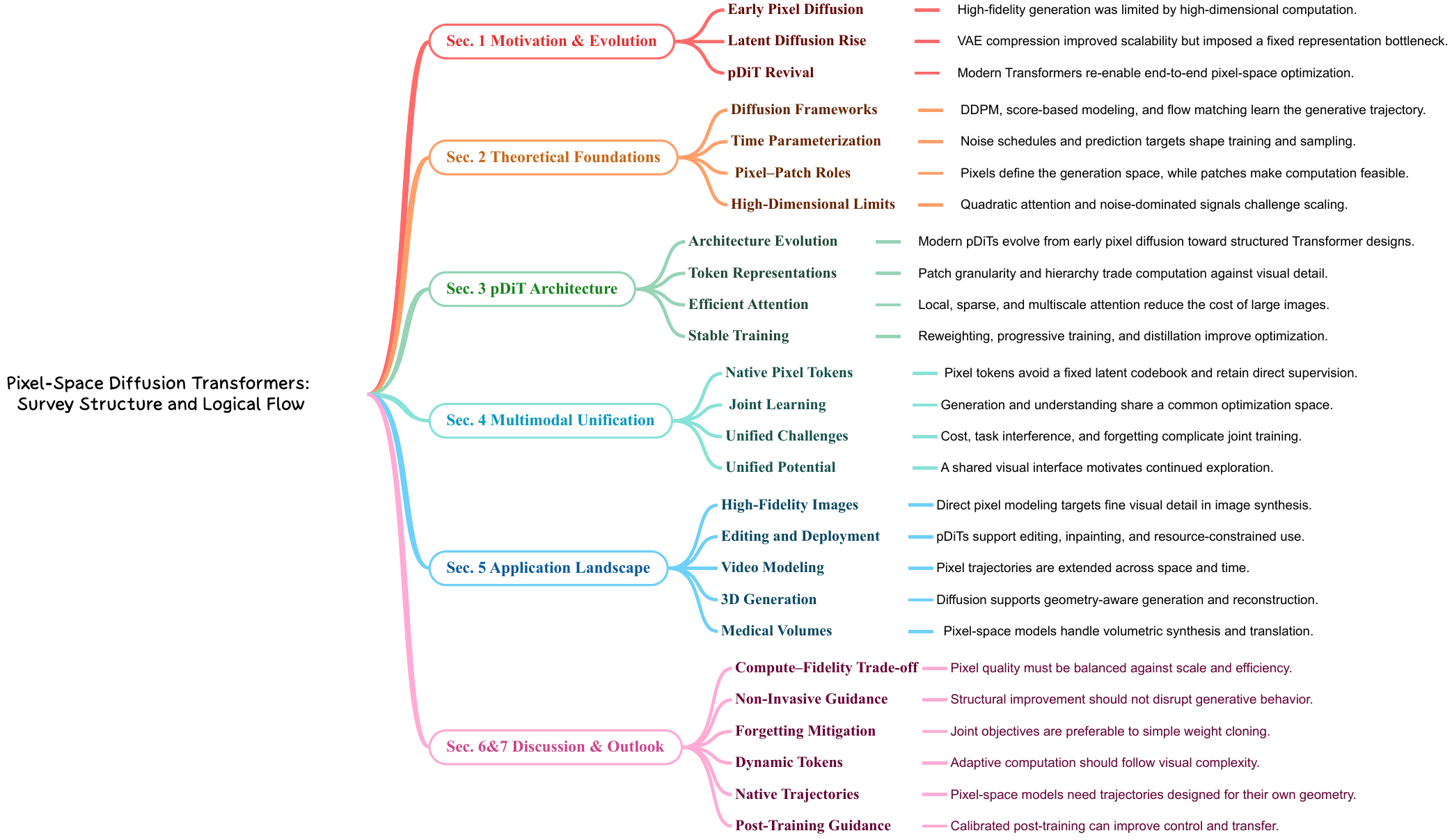}
    \caption{Overview of the Survey Structure and Research Roadmap.}
    \label{fig:framework}
\end{figure*}

Transitioning back to the pixel domain requires overcoming systemic optimization and computational challenges. Without a downsampled latent space, redesigning noise schedules, loss weighting schemes, and network scaling trajectories becomes paramount. Natural images typically reside on a low-dimensional manifold embedded within a vast, high-dimensional pixel space. To exploit this intrinsic structure, modern pixel-space frameworks employ a direct clean-image prediction ($x_0$-prediction) target, anchoring the network's optimization trajectories directly onto the data manifold and significantly accelerating convergence~\cite{li2026back}.
Simultaneously, to manage the computational load while preserving details across full-resolution canvases, recent breakthroughs have introduced multi-branch architectures~\cite{chen2025pixelflow}, frequency-decoupling mechanisms~\cite{ma2026deco}, and adaptive multi-scale tokenization strategies~\cite{yu2026pixeldit}. These innovations allow the network to handle global scene structures and microscopic textures within dedicated components, demonstrating that the pixel-space revival is driven not just by raw hardware scaling, but by targeted algorithmic solutions to high-dimensional visual modeling.

The revival of pixel-space diffusion is also profoundly reshaping the construction paradigm of unified multimodal foundation models that integrate understanding and generation~\cite{cai2026hidream,xie2025show,zhou2025transfusion}. 
Historically, unified models have suffered from a severe granularity schism: visual understanding demands highly abstract, invariant representations that disregard local pixel perturbations, whereas visual generation requires the exact opposite—absolute sensitivity to spatial layouts, color gradients, and fine-grained structures. To reconcile this conflict, traditional systems are forced to employ fragmented, heterogeneous pipelines, relying on separate disjoint modules like VAEs for generation and CLIP~\cite{radford2021learning} or DINO~\cite{oquab2023dinov2} for understanding~\cite{rombach2022high}. A native pixel-space framework provides a new unified foundation for reconciling the representational conflict between visual understanding and visual generation.~\cite{hoogeboom2023simple,li2026back,yu2026pixeldit}. Instead of fixing image representations in advance with an independent visual encoder, it directly maps raw pixels, text tokens, and task conditions into a unified shared token space, where the same end-to-end Transformer module jointly learns high-level semantic abstractions and low-level pixel details~\cite{peebles2023scalable,hoogeboom2023simple,cai2026hidream}.

Under this unified pixel-space representation framework, diverse objectives, such as text-to-image generation, instruction-driven editing, and subject-personalized generation, are optimized simultaneously. The generative losses directly supervise image details like edge and texture precision, and the understanding or representation-alignment objectives can in turn shape the underlying visual tokens, thereby moving beyond the separation imposed by a fixed VAE latent space or a heterogeneous visual encoding space.  For complex tasks that simultaneously depend on high-level semantic reasoning and fine-grained pixel-level fidelity, including image editing~\cite{lu2026semantic}, interactive visual question answering, reference-conditioned generation, complex text rendering (OCR), precise local control, and multi-image contextual reasoning, this unified pixel-space representation demonstrates substantial potential.

To provide a comprehensive roadmap of this rapidly evolving paradigm and clarify its key technical evolution, this survey systematically categorizes and reviews the pixel-space diffusion literature across three fundamental dimensions:
\begin{itemize}
    \item Architectural Taxonomy of Latent-Free Transformers: We examine how modern architectures substitute the VAE compression layer by utilizing advanced patch-granularity designs, hierarchical topologies, global--local coordination blocks, cross-scale connections, and frequency-aware modeling to preserve scalability and fidelity.
    \item Generative Formulations and Dynamics: We deconstruct the formulation of diffusion and flow-matching processes operating directly on raw pixel states, analyzing how distinct time parameterizations, velocity/vector fields, and signal-to-noise evolution mechanics dictate the reconstruction of global vs. local visual components.
    \item Unified Multimodal and Multi-Task Networks: We explore how text, pixels, and task conditions interact within a shared token vocabulary, investigating how a unified Transformer balances the gradient demands of disparate tasks like language understanding, semantic reasoning, image editing, and pixel generation.

\end{itemize}

By covering model architectures, the mathematical foundations of diffusion and flow processes, and unified multimodal modeling, this paper provides a systematic overview of pixel-space diffusion. It further summarizes the key technical advances, current bottlenecks, and future directions toward unified vision foundation models. The overall framework of this survey is presented in Fig.~\ref{fig:framework}. \textbf{\textit{To the best of our knowledge, this is the first survey dedicated to the systematic review and comprehensive analysis of pixel-space diffusion models.}}

\section{Background and Theoretical Foundations}
\subsection{Background} 
\begin{figure*}
    \centering
    \begin{minipage}[t]{0.5\linewidth}
        \centering
        \includegraphics[width=\columnwidth]{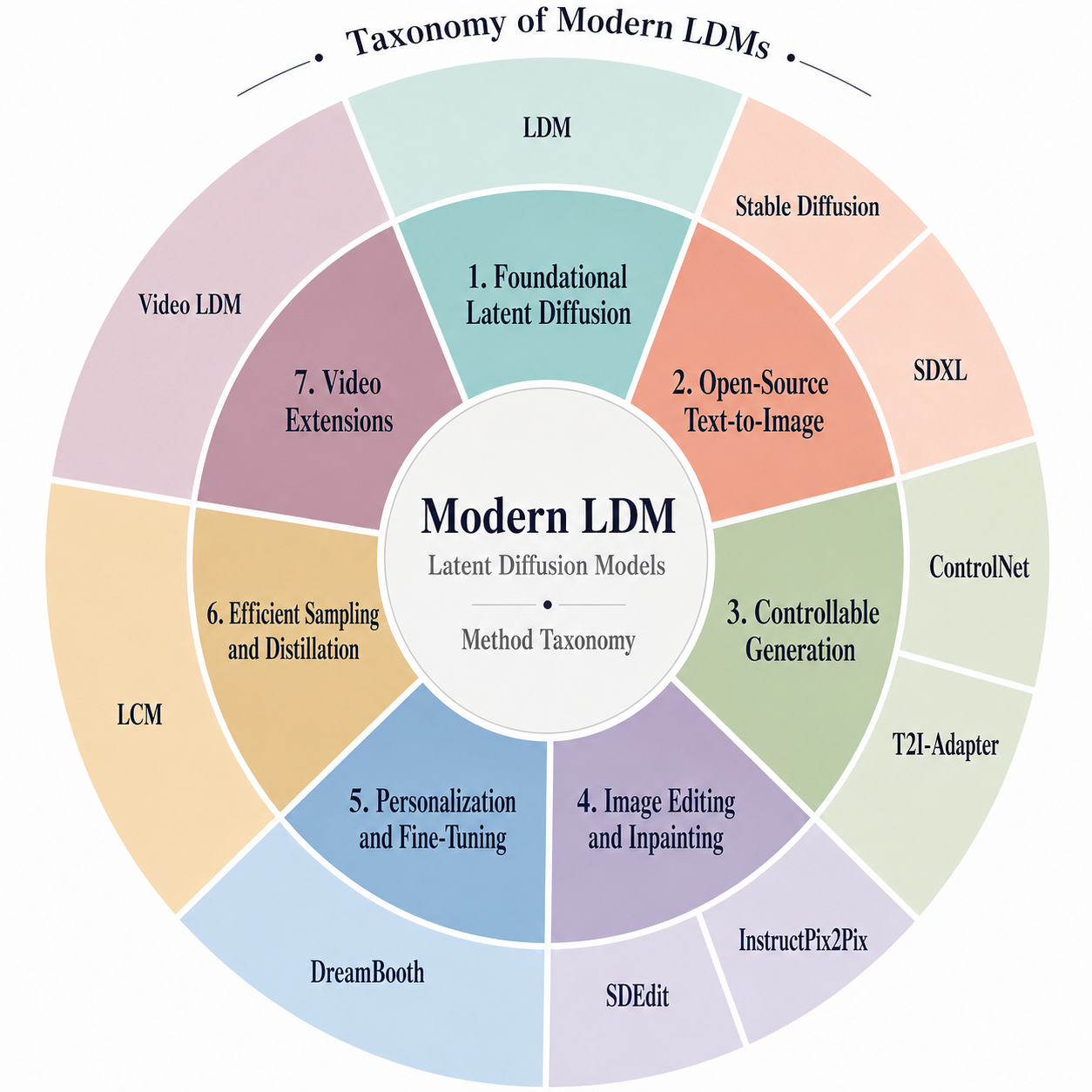}
    \end{minipage}
    \hfill
    \begin{minipage}[t]{0.49\linewidth}
        \centering
        \includegraphics[width=\columnwidth]{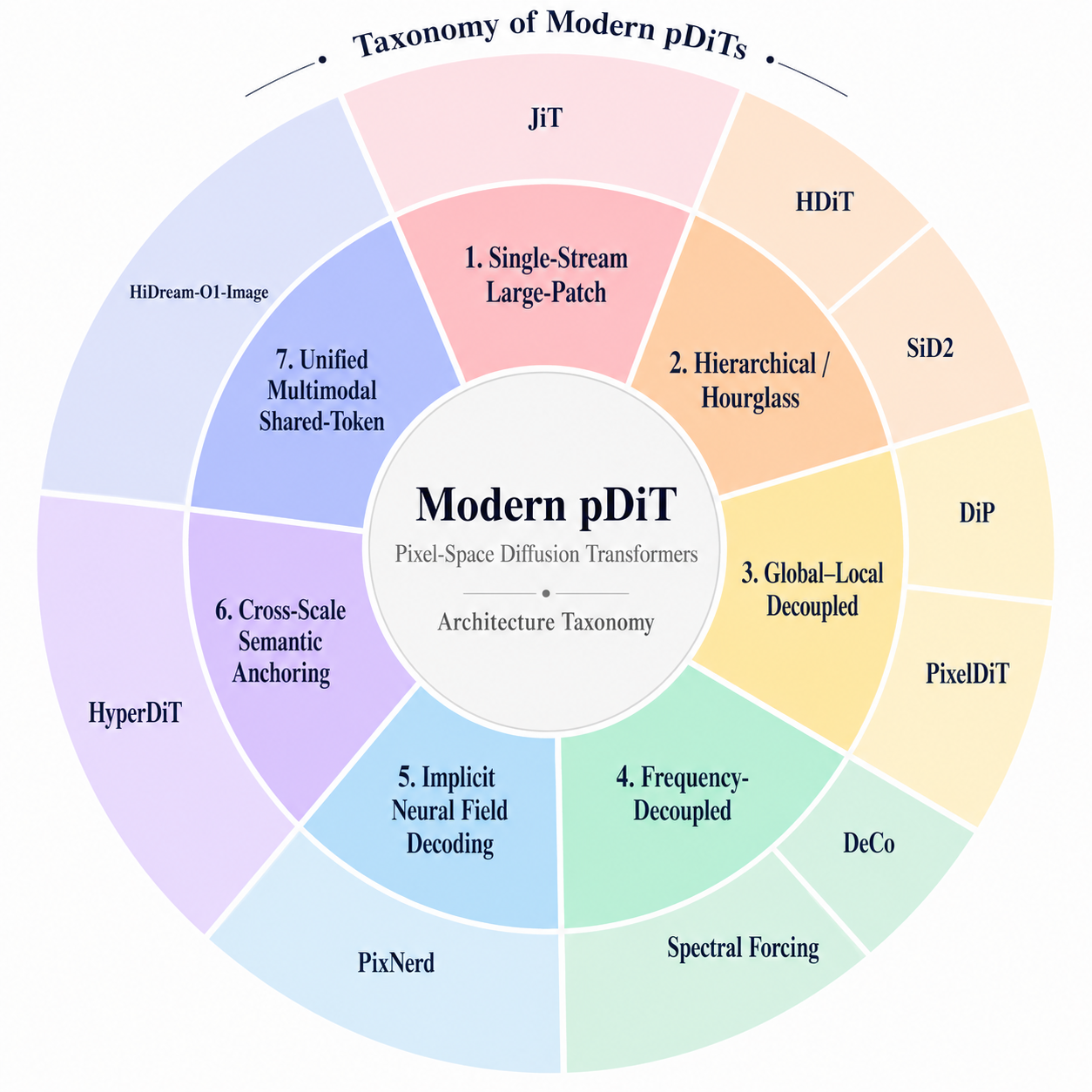}
    \end{minipage}
\caption{
    Taxonomy and representative methods of diffusion models:
    \textbf{Left:} Latent-space diffusion models (LDM);
    \textbf{Right:} Pixel-space diffusion models (pDiT).
}
    \label{fig:diffusion_taxonomy}
\end{figure*}

\noindent  \textit{\underline{The LDM Bottleneck}}:
Latent diffusion models (LDMs)~\cite{ciubotariu2026low,hong2026hyperspectral,bolya2026perception,moskalenko2026ntire,gushchin2026ntire,frans2025one,podell2024sdxl,blattmann2023align,lovelace2023latent,corneanu2024latentpaint} map high-dimensional images into a compact representation space through a pretrained VAE~\cite{kingma2013auto,zou2026turbo,li2026sparc3d,bi2026vision} or a VQ-VAE codebook~\cite{chen2025softvq,jia2025mgvq,guo2026improving,zhang2025tar3d}, and then learn the denoising process in that space. This design substantially reduces the number of tokens, memory consumption, and iterative generation cost, and therefore has established a prominent role in visual generation. However, these efficiency gains are built on lossy spatial compression: when the encoder maps images to low-dimensional continuous latents or discrete codebook indices, it must discard part of the pixel-level information, and the compressed or quantized content cannot be losslessly recovered from the latents by the subsequent diffusion model~\cite{he2026hyperdit,yanmultitune,yanintdiff}. Consequently, the upper bound of visual fidelity is determined not only by the diffusion backbone, but also by the spatial compression ratio of the VAE, the capacity of the latent variables, the expressive power of the encoder and decoder, and the representational preferences induced by the reconstruction objective~\cite{chen2026dip}. In practice, this representational bottleneck often manifests as smoothed micro-textures, distorted local geometry, degraded sharp edges, and artifacts when rendering small text, regular patterns, and high-frequency details~\cite{crowson2024scalable,ma2026deco}. More importantly, VAE/VQ modules are typically trained independently with reconstruction loss as the primary objective; their optimal compressed representations may not align with the requirements of downstream diffusion generation, controllable editing, or cross-modal conditional alignment, thereby further introducing a mismatch between the generative objective and the representation objective~\cite{chen2025pixelflow}. Fig.~\ref{fig:diffusion_taxonomy} (left) illustrates the architectures of LDMs and their representative methods.

\noindent \textit{\underline{Pixel-Space Diffusion Models}}:
Pixel-space diffusion models~\cite{zhang2024pixel,jain2023vectorfusion,baade2026latent,rozet2026lost} represent an important trend in which generative modeling moves from compressed latent modeling back to modeling in the original visual space. Compared with LDMs, which rely on a two-stage framework that first compresses images with a VAE and then performs generation, pixel-space diffusion directly takes the raw image as the diffusion object. This allows the generative loss to act on real pixel structures, thereby more naturally constraining edges, textures, colors, text, and fine-grained local structures~\cite{cheng2025leanvae}. This direction was long constrained by the computational burden of high-resolution images. Recent advances in pixel-space diffusion have shown that carefully designed noise schedules, loss weighting, and network architectures can substantially reduce this overhead, enabling end-to-end pixel diffusion to achieve strong performance with low computational cost in high-resolution image generation. After Simple Diffusion~\cite{hoogeboom2023simple}, works such as JiT~\cite{li2026back} and PixelDiT~\cite{yu2026pixeldit} further combined pixel-space diffusion with Transformer architectures, exploring scalable generation without relying on VAEs from the perspectives of simple large-patch image Transformers and dual-layer/end-to-end pixel DiT designs, respectively. These works further explore how to enable direct pixel-space modeling while alleviating the computational and scalability bottlenecks of conventional pixel-space diffusion models. Furthermore, HiDream-O1-Image~\cite{cai2026hidream} maps raw pixels, text tokens, and task-specific conditions into a shared token space, demonstrating the potential of pixel-level unified Transformers for tasks such as text-to-image generation, instruction-based editing, and subject personalization. Thus, the significance of pixel-space diffusion extends beyond merely improving image details; it also provides a more unified visual foundation for the next generation of multimodal models that integrate understanding and generation. The major architectures of pDiTs and their representative methods are illustrated in Fig.~\ref{fig:diffusion_taxonomy} (right).

\subsection{Theoretical Foundations}

Pixel-space generative models are grounded in two paradigms: stochastic denoising frameworks, represented by Denoising Diffusion Probabilistic Models (DDPMs)~\cite{ho2020denoising}, and deterministic continuous-flow frameworks, represented by Flow Matching (FM)~\cite{lipman2022flow} and Rectified Flow (RF)~\cite{liu2022flow}. Although both paradigms aim to learn a mapping from a noise distribution to the data distribution, they differ fundamentally in how the trajectory is parameterized.

\noindent \textit{\underline{(1) Denoising Diffusion Probabilistic Models}}: DDPM formulates image generation as a discrete-time Markov reverse-diffusion process. Starting from Gaussian noise, the model progressively removes perturbations through a sequence of denoising state transitions and eventually recovers a clean image. Its forward diffusion process is typically defined as:
\begin{equation}
q(\mathbf{x}_t \mid \mathbf{x}_{t-1})
=
\mathcal{N}
\left(
\sqrt{1-\beta_t}\,\mathbf{x}_{t-1},
\beta_t\mathbf{I}
\right).
\end{equation}
$q(\mathbf{x}_t \mid \mathbf{x}_{t-1})$ denotes the transition distribution from $\mathbf{x}_{t-1}$ to $\mathbf{x}_t$ in the forward process, where $\mathbf{x}_{t-1}$ and $\mathbf{x}_t$ are the latents at diffusion time steps $t-1$ and $t$, respectively.
\(\mathcal{N}(\mu,\Sigma)\) denotes a Gaussian distribution with mean \(\mu\) and covariance \(\Sigma\). \(\beta_t\in(0,1)\) is the predefined noise-scheduling coefficient at time step \(t\), which controls the intensity of the noise added at the current step. \(\mathbf{I}\) denotes the identity matrix. This transition gradually injects Gaussian noise into \(\mathbf{x}_{t-1}\) while retaining part of the original signal.
Through recursive derivation, the sampling form at an arbitrary time step can be obtained directly:
\begin{equation}
\mathbf{x}_t
=
\sqrt{\bar{\alpha}_t}\,\mathbf{x}_0
+
\sqrt{1-\bar{\alpha}_t}\,\epsilon,
\qquad
\epsilon\sim\mathcal{N}(0,\mathbf{I}).
\end{equation}
\(\mathbf{x}_0\) denotes the clean image. \(\epsilon\) denotes random noise sampled from the standard Gaussian distribution \(\mathcal{N}(0,\mathbf{I})\). \(\alpha_t\) is the signal-retention coefficient at time step \(t\), complementary to the noise-scheduling coefficient \(\beta_t\), with $\alpha_t=1-\beta_t$. \(\bar{\alpha}_t\) denotes the cumulative signal-retention coefficient from time step \(1\) to time step \(t\), defined as $\bar{\alpha}_t
=
\prod_{s=1}^{t}\alpha_s$, where \(s\) is the time-step index in the product. As \(t\) increases, \(\bar{\alpha}_t\) typically decreases, indicating that the original image signal continuously decays while the noise component progressively increases. In this expression, \(\sqrt{\bar{\alpha}_t}\) controls the proportion of the clean image signal retained in \(\mathbf{x}_t\), whereas \(\sqrt{1-\bar{\alpha}_t}\) controls the strength of the injected Gaussian noise. The equation enables direct sampling of \(\mathbf{x}_t\) at any time step from the clean image \(\mathbf{x}_0\), without executing the full forward diffusion chain step by step.
During training, DDPM typically adopts a noise-prediction objective, whose standard loss function is formulated as:
\begin{equation}
\mathcal{L}_{\mathrm{DDPM}}
=
\mathbb{E}_{x_0,\epsilon,t}
\left[
\left\|
\epsilon-
\epsilon_\theta(\mathbf{x}_t,t)
\right\|_2^2
\right].
\end{equation}
\(\mathcal{L}_{\mathrm{DDPM}}\) denotes the noise-prediction loss. The expectation \(\mathbb{E}_{\mathbf{x}_0,\epsilon,t}[\cdot]\) is taken with respect to \(\mathbf{x}_0\),  \(\epsilon\), and \(t\). \(\epsilon_\theta(\mathbf{x}_t,t)\) denotes the noise predicted by the denoising network. \(\|\cdot\|_2^2\) denotes the squared \(L_2\) norm. By minimizing the mean squared error between the true noise and the predicted noise, this objective enables the model to learn the reverse denoising process.

\noindent\textit{\underline{(2) Flow Matching}}: 
Unlike DDPM, which progressively estimates conditional probabilities through a discrete-time reverse-diffusion chain, FM
directly learns a time-dependent velocity field that transports samples along continuous trajectories from a simple source distribution to the target image distribution. Let the source distribution be a standard Gaussian distribution:
\begin{equation}
p_0(\mathbf{x})
=
\mathcal{N}
\left(
\mathbf{x};\mathbf{0},\mathbf{I}
\right).
\end{equation}
\(p_0(\mathbf{x})\) denotes the source probability distribution at the initial time \(t=0\) of the continuous flow. \(\mathbf{x}\) denotes a random variable in this distribution. \(\mathbf{0}\) and \(\mathbf{I}\) denote the zero-mean vector and the identity covariance matrix, respectively. Thus, the source distribution is typically set to an easily sampled standard Gaussian distribution.
\begin{figure*}[t]
    \centering
    \includegraphics[width=1.0\linewidth]{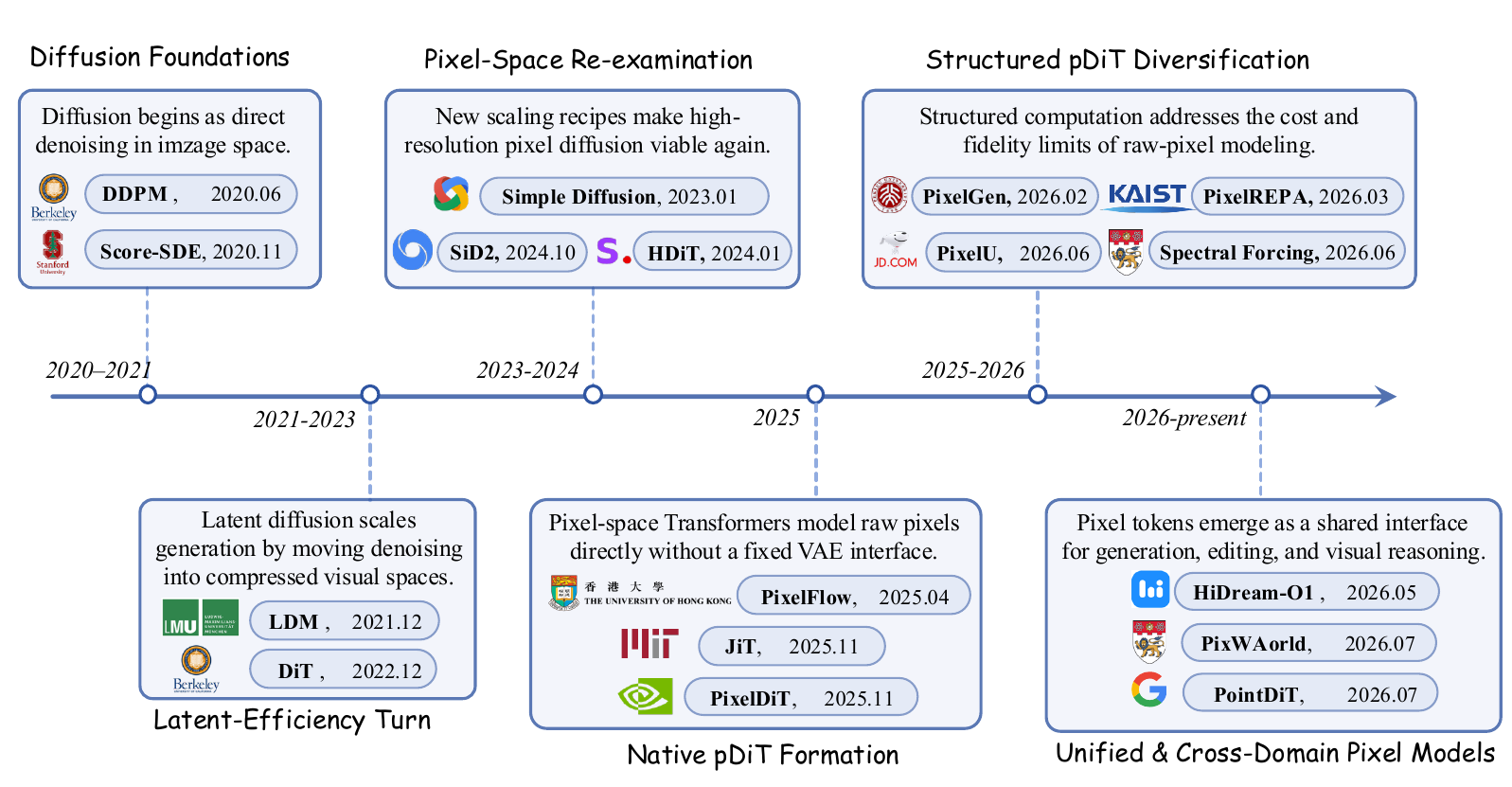}
    \caption{Timeline of the development of pixel diffusion models.}
    \label{fig:timeline}
\end{figure*}
The target distribution is the real image data distribution, $p_1(\mathbf{x}) = p_{\mathrm{data}}(\mathbf{x})$. The objective of FM is to learn a continuous velocity field that transports the source distribution $p_0$ to the target distribution $p_1$. Here, $t \in [0,1]$ denotes the normalized flow time, with $t = 0$ corresponding to the source noise distribution and $t = 1$ to the target image distribution. Accordingly, in the Flow Matching formulation, $\mathbf{x}_0$ denotes the initial noise, whereas $\mathbf{x}_1$ denotes the clean image. This differs from the notation used above for DDPM, where $\mathbf{x}_0$ denotes the clean image.
A commonly used conditional probability path is the linear interpolation path connecting the source and target samples:
\begin{equation}
\label{flow_xt}
\mathbf{x}_t
=
(1-t)\mathbf{x}_0
+
t \mathbf{x}_1.
\end{equation}
This equation defines a linear conditional probability path connecting \(\mathbf{x}_0\) and \(\mathbf{x}_1\): as \(t\) increases from \(0\) to \(1\), \(\mathbf{x}_t\) gradually transitions from pure noise to a clean image.
Under this path, the conditional velocity field is:
\begin{equation}
\mathbf{u}_t
\left(
\mathbf{x}_t\mid \mathbf{x}_0,\mathbf{x}_1
\right)
=
\frac{\mathrm{d}\mathbf{x}_t}{\mathrm{d}t}
=
\mathbf{x}_1-\mathbf{x}_0.
\end{equation}
\(\mathbf{u}_t(\mathbf{x}_t\mid \mathbf{x}_0,\mathbf{x}_1)\) denotes the conditional velocity field at time \(t\) for latent \(\mathbf{x}_t\). \(\mathrm{d}\mathbf{x}_t/\mathrm{d}t\) denotes the derivative of the generative latent with respect to continuous time \(t\). \(\mathbf{x}_1-\mathbf{x}_0\) denotes the transport direction from the source sample to the target sample.
FM aims to train a parameterized velocity field \(\mathbf{v}_{\theta}(\mathbf{x}_t,t)\) to approximate the conditional velocity above. The corresponding conditional flow matching loss can be written as:
\begin{equation}
\mathcal{L}_{\mathrm{CFM}}(\theta)
=
\mathbb{E}_{
t\sim\mathcal{U}(0,1),
\,\mathbf{x}_0\sim p_0,
\,\mathbf{x}_1\sim p_1
}
\left[
\left\|
\mathbf{v}_{\theta}(\mathbf{x}_t,t)
-
\left(\mathbf{x}_1-\mathbf{x}_0\right)
\right\|_2^2
\right].
\end{equation}
\(\mathcal{U}(0,1)\) denotes the uniform distribution over the interval \([0,1]\).
\(\|\cdot\|_2^2\) denotes the squared \(L_2\) norm. By regressing the conditional velocity, this loss trains the model to learn the continuous transport direction from the noise distribution toward the real image distribution.
Although the conditional velocity field defined by specific endpoint samples is used during training, the velocity field ultimately learned by the model corresponds to the effective transport direction along the marginal probability path \(p_t(\mathbf{x})\). It can be expressed as:
\begin{equation}
\mathbf{v}_t^{\ast}(\mathbf{x})
=
\mathbb{E}
\left[
\mathbf{x}_1-\mathbf{x}_0
\mid
\mathbf{x}_t=\mathbf{x}
\right].
\end{equation}
$\mathbf{v}_t^{\ast}(\mathbf{x})$ denotes the optimal velocity field at time $t$ evaluated at a latent state $\mathbf{x}$ along the marginal probability path $p_t(\mathbf{x})$. \(\mathbb{E}[\cdot\mid \mathbf{x}_t=\mathbf{x}]\) denotes the conditional expectation over the velocities associated with all possible endpoint sample pairs, conditioned on \(\mathbf{x}_t=\mathbf{x}\). This expression shows that, although the training supervision is constructed from specific endpoint pairs, the model ultimately learns the average of all possible conditional transport directions given the intermediate latent.

Rectified Flow (RF) further emphasizes the straightening of transport trajectories on the basis of Flow Matching. Its core objective is to reduce unnecessary curvature in the paths between the source and target distributions, making the velocity field learned by the model more consistent across different time steps. After iterative reflow, the generation trajectories can progressively approach linear paths, thereby reducing numerical integration error and improving the stability of few-step sampling. For pixel-space Diffusion Transformers (pDiTs), continuous flow modeling provides a mathematical perspective distinct from traditional discrete denoising: the model no longer merely predicts a local residual at a particular noise level, but directly learns the continuous transport direction from high-dimensional noisy images to the natural image manifold. Consequently, Flow Matching and Rectified Flow are gradually becoming important theoretical foundations for efficient sampling, direct image prediction, and end-to-end pixel generation in modern pDiTs models.

\section{Pixel-Space Diffusion Transformer}

In Pixel-Space Diffusion Transformers (pDiTs), the generative process operates directly in the image space, without relying on a pretrained VAE or VQ codebook as a fixed image-compression interface. Within this pixel-space formulation, the Transformer~\cite{vaswani2017attention} serves as the primary backbone, operating on tokenized image representations through attention mechanisms and enabling end-to-end generative modeling in pixel space. The evolution of pDiTs is illustrated in Fig.~\ref{fig:timeline}. More specifically, given a pixel-space state $\mathbf{x}_t \in \mathbb{R}^{H \times W \times C}$, a pDiTs typically first partitions it into local patches of size $P \times P$, and then obtains visual tokens through a learnable mapping:
\begin{equation}
\mathbf{z}_t =
\mathcal{E}_{\theta}
\left(
\mathrm{Patchify}_{P}(\mathbf{x}_t)
\right).
\end{equation}
$H$, $W$, and $C$ denote the image height, width, channel count, respectively. $P$ denotes the side length of each local patch, and $\mathrm{Patchify}_{P}(\cdot)$ denotes the operation that partitions the input image into non-overlapping local patches of size $P \times P$. $\mathcal{E}_{\theta}(\cdot)$ denotes a learnable patch-embedding module parameterized by $\theta$, which maps local pixel patches into visual tokens. $\mathbf{z}_t$ denotes the visual token sequence obtained from the pixel-space state $\mathbf{x}_t$. Subsequently, the Transformer backbone models the relationships among tokens conditioned on $t$ and an optional text condition $\mathbf{c}$, and predicts the generation target through an end-to-end prediction head:
\begin{equation}
\hat{\mathbf{y}}_t =
\mathcal{D}_{\theta}
\left(
\mathcal{T}_{\theta}
(\mathbf{z}_t,t,\mathbf{c})
\right).
\end{equation}
Here, $\mathcal{T}_{\theta}(\cdot)$ denotes the Transformer backbone parameterized by $\theta$, which models global dependencies among tokens given the visual token sequence $\mathbf{z}_t$, the diffusion time step $t$, and the optional text condition $\mathbf{c}$. $\mathcal{D}_{\theta}(\cdot)$ denotes the end-to-end pixel decoding head parameterized by $\theta$, which maps the hidden representations produced by the Transformer to the pixel-space prediction target. $\hat{\mathbf{y}}_t$ denotes the predicted generation target at time step $t$, which may correspond to noise, velocity, or the clean image depending on the prediction parameterization.



\subsection{From Early Pixel Diffusion to Modern pDiTs}

\noindent \textbf{Why Was Early Pixel-Space Diffusion Not Mainstream?}

Early DDPMs~\cite{ho2020denoising}, score-based models~\cite{song2019generative}, and SDE-based models~\cite{song2020score} typically performed forward noising and reverse denoising directly in the pixel space. This paradigm has an end-to-end advantage, because the model can learn the image distribution directly without additional encoding and decoding interfaces. However, as image resolution increases, the limitations of modeling in raw pixel space become increasingly apparent. 
These limitations are reflected in three aspects: (1) the computational and memory burden induced by repeated denoising at high resolutions, (2) the optimization difficulty caused by the high-dimensional noisy pixel space, and (3) the limited scalability of predominantly convolutional backbones for modeling complex long-range semantic relationships.

\noindent \underline{\textit{(1) Computational Bottleneck}}. Pixel-space models take full-resolution images as input and repeatedly process high-resolution spatial representations during each denoising step. For denoising networks based on convolutional U-Net architectures~\cite{si2024freeu,tian2026u,ronneberger2015u,trebing2021smaat,he2022swin}, although the per-layer computation does not exhibit the quadratic complexity of global attention, dense convolutions together with feature downsampling and upsampling are still performed across multiple resolution levels. Accordingly, the computational cost of pixel-space diffusion sampling can be approximated as:
\begin{equation}
\mathcal{C}_{\mathrm{pixel}}
\propto
N_{\mathrm{FE}}
\sum_{l=1}^{L}
H_l W_l k_l^2
C_l^{\mathrm{in}} C_l^{\mathrm{out}},
\end{equation}
where $N_{\mathrm{FE}}$ denotes the number of network evaluations during diffusion sampling, $H_l$ and $W_l$ denote the spatial resolution at the $l$-th level, $k_l$ is the convolution kernel size, and $C_l^{\mathrm{in}}$ and $C_l^{\mathrm{out}}$ denote the numbers of input and output channels, respectively. As the input image resolution increases, high-resolution feature maps substantially increase both the computational cost and peak memory consumption of each network evaluation. Since diffusion sampling requires repeated network evaluations over multiple denoising steps, the overall computational cost further accumulates along the sampling trajectory, making direct pixel-space generation difficult to scale efficiently to high-resolution settings~\cite{rombach2022high,peebles2023scalable}.

\noindent\underline{\textit{(2) Optimization Difficulty}}. Early pixel-space models faced a conflict between learnability and fidelity. Raw pixel space preserves rich local textures, edge structures, and fine-grained color variations, and therefore in principle offers a higher upper bound on visual fidelity. However, high-dimensional pixel variables also require the model to simultaneously learn low-frequency global semantics, high-frequency visual details, and the complex data geometry associated with different noise levels~\cite{dhariwal2021diffusion,rombach2022high,hoogeboom2023simple}. In particular, under the traditional diffusion training paradigm dominated by noise prediction, the model must recover a complex target in a high-dimensional noisy space, which often slows the formation of semantic structure. The model may preferentially fit local pixel statistics while struggling to quickly establish stable object relations, spatial layouts, and long-range semantic dependencies.

\noindent\underline{\textit{(3) Long-Range Modeling}}.  In addition, early pixel-space diffusion models typically used convolutional U-Nets as the denoising backbone. Such networks mainly rely on local convolutions to gradually enlarge the receptive field, and aggregate long-range information through repeated downsampling and bottleneck layers. As a result, relationships between spatially distant regions in an image cannot be modeled directly; they must instead be established indirectly through multiple layers of feature propagation. For complex text-conditioned generation, the model often needs to coordinate multiple objects, attribute bindings, relative positions, and scene-level layouts involving long-range dependencies. However, convolutional backbones model such global relationships primarily through bottleneck features at limited resolution, which can easily lead to unstable object relations, attribute mismatches, or inconsistencies in the overall composition.

\noindent \textbf{Why May Vision Foundation Models Not Work as Expected in Diffusion Models?}

Vision Foundation Models (VFMs), such as CLIP~\cite{radford2021learning}, DINOv2~\cite{oquab2023dinov2}, and DINOv3~\cite{simeoni2025dinov3}, have learned visual representations with strong semantic discriminability, object-level consistency, and a certain degree of spatial organization. Introducing these representations into diffusion models can alleviate the slow convergence that arises when high-level semantic structures are learned solely from the denoising objective. However, discriminative visual representations and generative visual representations are optimized for different objectives. The former tend to compress local variations irrelevant to recognition tasks in order to obtain semantic representations that are more stable to scale, texture, color, and local perturbations; the latter must preserve these variations as much as possible in order to recover precise edges, textures, colors, text, and fine-grained spatial structures. Therefore, the representations provided by VFMs are better viewed as auxiliary semantic priors rather than as complete visual representations required by generative models. Based on this distinction, existing methods generally introduce VFM representations into diffusion models as auxiliary priors rather than directly treating them as complete generative representations. We next examine how such priors are incorporated through representation alignment and VFM-based visual tokenization, as well as the limitations of these strategies.

\noindent \underline{\textit{(1) Representation Alignment for Diffusion Models.}}
Representation Alignment (REPA)~\cite{yu2024representation} is a representative approach for introducing VFM priors into the diffusion backbone. Its basic idea is to use a frozen visual encoder \(E_{\phi}\) to extract teacher features from the clean image \(\mathbf{x}_0\), and to constrain the diffusion Transformer to produce intermediate representations close to those features when processing the noisy input \(\mathbf{x}_t\). Let the hidden state of the \(l\)-th Transformer block in the diffusion model be:
\begin{equation}  
\mathbf{H}^{(l)}_{\theta}(x_t,t)
=
\left[
\mathbf{h}^{(l)}_1,\ldots,\mathbf{h}^{(l)}_N
\right],
\end{equation}
and let the patch-level representations from the teacher be:
\begin{equation}
\mathbf{R}_{\phi}(\mathbf{x}_0)
=
\left[
\mathbf{r}_1,\ldots,\mathbf{r}_N
\right]
=
E_{\phi}(\mathbf{x}_0),
\end{equation}
where \(N\) denotes the number of visual tokens. Because the feature dimensions of the diffusion backbone and the teacher model are typically different, REPA uses a learnable projection \(g_{\psi}\) to map diffusion features into the teacher feature space:
\begin{equation}
\hat{\mathbf{r}}_i
=
g_{\psi}\left(\mathbf{h}^{(l)}_i\right).
\end{equation}
The token-wise alignment loss can be written as:
\begin{equation}
\mathcal{L}_{\mathrm{REPA}}
=
\frac{1}{N}
\sum_{i=1}^{N}
\left[
1-
\frac{
\hat{\mathbf{r}}_i^{\top}\mathbf{r}_i
}{
\left\|\hat{\mathbf{r}}_i\right\|_2
\left\|\mathbf{r}_i\right\|_2
}
\right],
\end{equation}
and the complete training objective is therefore:
\begin{equation}
\mathcal{L}
=
\mathcal{L}_{\mathrm{gen}}
+
\lambda_{\mathrm{REPA}}
\mathcal{L}_{\mathrm{REPA}},
\end{equation}
where \(\mathcal{L}_{\mathrm{gen}}\) denotes the diffusion or flow-matching objective, and \(\lambda_{\mathrm{REPA}}\) controls the trade-off between generative learning and representation alignment. During training, the VFM parameters \(\phi\) remain frozen, and only the diffusion backbone and the projection head are updated. The alignment branch can be removed at inference time and therefore introduces no additional sampling cost.
The main role of REPA is to reduce the optimization difficulty of learning high-level semantic structures in diffusion models. By introducing semantic representations of clean images into the front part of the network, the model can establish object categories, overall layouts, and global relations earlier, rather than relying entirely on the denoising loss to develop these abilities from scratch. Existing experiments have demonstrated that this mechanism can significantly accelerate the training convergence of DiT~\cite{peebles2023scalable} and SiT~\cite{ma2024sit}, and improve generation metrics under limited training budgets. However, these results mainly indicate that VFM priors can provide better semantic initialization and optimization trajectories; they are not sufficient to prove that such priors can improve the final modeling upper bound of generative models for high-frequency textures, fine edges, text, regular geometric structures, or local color variations. 

In other words, faster semantic convergence does not necessarily imply improved pixel-level fidelity. Specifically, this limitation stems from the inherent multi-objective conflict in representation alignment. On the one hand, the diffusion model must minimize the generative loss to learn the complete data distribution; on the other hand, it must approximate the discriminative representation space defined by a frozen VFM:
\begin{equation}
\min_{\theta,\psi}
\quad
\mathcal{L}_{\mathrm{gen}}
+
\lambda_{\mathrm{align}}\mathcal{L}_{\mathrm{align}}.
\end{equation}
When the gradient directions of the two objectives are inconsistent, alignment supervision may accelerate the formation of semantic structure while simultaneously suppressing the fine-grained information required by the generative objective. In particular, in the later stages of training, when the diffusion model begins to learn visual variations that are actively compressed or ignored by the teacher representation, continuously imposing the alignment constraint may shift from an early optimization prior into a regularization bottleneck that limits the model's generative capacity. This also explains why subsequent studies have observed that the benefits of REPA may saturate or even decline in later training stages, motivating early termination of alignment, stage-wise training, or more flexible relation-level supervision.
This conflict is more pronounced in pixel space. Pixel diffusion models directly model high-dimensional, information-complete image variables, whereas teachers such as DINO or CLIP typically compress images into low-resolution and invariant semantic tokens. Directly requiring high-dimensional generative features to regress token by token to low-dimensional semantic targets creates a clear information asymmetry: many images with the same teacher representation but different textures, backgrounds, colors, or local shapes may be mapped to similar supervision targets. In this case, the model can reduce the alignment loss by preferentially recovering the category-level semantics retained by the teacher, without learning the complete pixel distribution, thereby forming an alignment shortcut. PixelREPA~\cite{shin2026representation}, in its study of JiT~\cite{li2026back}, further shows that standard REPA in pixel space may cause FID~\cite{heusel2017gans} to deteriorate in later training stages and reduce generation diversity within semantically similar image subsets. Therefore, token-wise alignment strategies that are effective in latent space cannot be transferred unconditionally to pixel-space models.

\noindent\underline{\textit{(2) VFM-based Tokenizers and the Reconstruction Ceiling.}}
RAE~\cite{zheng2025diffusion}, VA-VAE~\cite{yao2025reconstruction,leng2025repa}, and VFM-VAE~\cite{bi2026vision} introduce vision foundation models from another perspective: rather than merely aligning intermediate features of the diffusion backbone, they directly use or distill VFM representations to construct the tokenizer of the generative model. RAE combines frozen DINO, SigLIP~\cite{zhai2023sigmoid}, or MAE~\cite{he2022masked} encoders with a trainable decoder to form a representation autoencoder. VA-VAE and VFM-VAE seek a better balance among reconstruction quality, semantic organization capability, and downstream diffusion learnability through semantic feature constraints or specially designed decoders. These methods show that semantically richer latents can significantly improve the optimization efficiency of diffusion Transformers and expand the reconstruction--generation Pareto frontier of traditional VAEs~\cite{sener2018multi,lin2019pareto,kaplan2020scaling,hoffmann2022training,tan2019efficientnet}.
However, such methods still cannot completely eliminate the information bottleneck introduced by fixed visual encoders. If the encoder is invariant to fine-grained textures, color variations, small-scale text, or precise geometric relations, such information may already have been compressed before entering the diffusion model. Even if the subsequent decoder has strong reconstruction capability, it can only infer missing details from the retained representation and cannot guarantee per-sample recovery of the original information. Therefore, VFM-based tokenizers can improve the semantic quality and empirical reconstruction performance of traditional tokenizers, but their reconstruction upper bound remains constrained by the information sufficiency of the fixed encoded representation. It should be emphasized that this ``reconstruction ceiling'' problem mainly applies to tokenizer-level methods such as RAE, VA-VAE, and VFM-VAE, and should not be directly used to describe REPA, which only adds an auxiliary loss during training.

\noindent \underline{\textit{(3) From Feature Matching to Structural Alignment.}}
The above limitations have driven a shift from direct feature matching to relation-level structural constraints. Spatial Gram Alignment (SGA)~\cite{zhang2026spatial} no longer forces generative features to replicate the specific channel representations of a VFM, but instead aligns self-similarity relations among different spatial locations. Let the generated features and VFM features be denoted respectively as:
\begin{equation}
\mathbf{F}_{\mathrm{gen}}\in\mathbb{R}^{C_g\times N},
\qquad
\mathbf{F}_{\mathrm{VFM}}\in\mathbb{R}^{C_v\times N},
\end{equation}
and their normalized Spatial Gram matrices are expressed as:
\begin{equation}
\mathbf{G}_{\mathrm{gen}}
=
\bar{\mathbf{F}}_{\mathrm{gen}}^{\top}
\bar{\mathbf{F}}_{\mathrm{gen}},
\qquad
\mathbf{G}_{\mathrm{VFM}}
=
\bar{\mathbf{F}}_{\mathrm{VFM}}^{\top}
\bar{\mathbf{F}}_{\mathrm{VFM}},
\end{equation}
with the corresponding structural alignment objective:
\begin{equation}
\mathcal{L}_{\mathrm{SGA}}
=
\left\|
\mathbf{G}_{\mathrm{gen}}
-
\mathbf{G}_{\mathrm{VFM}}
\right\|_{F}^{2}.
\end{equation}
Because this objective transfers only relative relations among spatial locations, without requiring two models to have identical feature values or channel semantics, SGA can alleviate the conflict between discriminative and generative representations to some extent: VFM provides priors for macroscopic object relations and spatial organization, while the generative objective continues to recover microscopic textures and pixel details. Nevertheless, SGA remains a form of auxiliary structural supervision rather than a fundamental solution to the information-loss problem in VFM representations.

\noindent \underline{\textit{Overall.}} Vision foundation models such as DINO and CLIP can improve the semantic learning efficiency of diffusion models, but they should not be regarded as substitutes for solving generative modeling. REPA-style methods accelerate the semantic convergence of the diffusion backbone; RAE, VA-VAE, and VFM-VAE attempt to construct latents that are both semantic and reconstructible; and SGA reduces perturbations to native generative representations by relaxing the form of alignment. Together, these methods reveal a principle: discriminative visual priors can help generative models determine what should appear in an image and how it should be organized, but high-fidelity generation still requires the model to preserve and learn low-level visual information that VFMs may ignore. For pixel-space diffusion, the advantage lies in avoiding the use of fixed, compressed, and task-specific external representations as a hard upper bound on generative information, thereby allowing semantic structure and pixel details to emerge jointly under a unified end-to-end objective.

\noindent \textbf{Why Are Pixel-Space Diffusion Models Re-emerging?}

Recently, pixel-space diffusion has regained attention~\cite{hoogeboom2025simpler,li2026back} not merely because of increased computational resources, but because training objectives, architectural design, representation supervision, and scalability have changed together.

First, modern training recipes have significantly improved the optimization stability of high-resolution pixel diffusion. Simple Diffusion shows that, by readjusting noise schedules, loss weights, network scaling strategies, and high-resolution feature processing, end-to-end pixel diffusion can substantially narrow the gap with latent-space methods~\cite{hoogeboom2023simple}. Subsequent studies further point out that the performance bottleneck in pixel space does not arise entirely from resolution itself, but is closely related to how the model allocates capacity to different frequency components and samples of different difficulty~\cite{ma2026deco}.

Second, predicting the clean image provides a more suitable learning objective for high-dimensional pixel trajectories~\cite{li2026back}. JiT proposes to directly predict $\hat{x}_0$, and then recover the velocity estimate from the predicted image. This method re-anchors the network output to the low-dimensional data manifold on which natural images lie, rather than directly regressing to noise or velocity targets distributed throughout the high-dimensional space. As a result, the model can form image structures earlier, while enabling large-patch Transformers to remain learnable in the original pixel space.
\begin{table*}[t]

\centering
\begingroup

\renewcommand{\CenterCell}[1]{%
  \begin{minipage}[c]{\linewidth}
  \centering
  \vspace*{1pt}
  #1\par
  \vspace*{1pt}
  \end{minipage}%
}
\renewcommand{\LeftCell}[1]{%
  \begin{minipage}[c]{\linewidth}
  \raggedright
  \setlength{\parindent}{0pt}%
  \vspace*{1pt}
  \noindent\ignorespaces#1\unskip\par
  \vspace*{1pt}
  \end{minipage}%
}
\newcommand{\CategoryCell}[1]{%
  \begin{minipage}[c]{\linewidth}
  \raggedright
  \setlength{\parindent}{0pt}%
  \vspace*{1pt}
  \noindent\ignorespaces#1\unskip\par
  \vspace*{1pt}
  \end{minipage}%
}

{\Large
\bfseries
\textcolor{TitleBlue}{Overview of Modern pDiT Architecture Taxonomy}
}

\vspace{0.25em}

\fontsize{8.3pt}{9.8pt}\selectfont
\renewcommand{\arraystretch}{0.95}
\setlength{\tabcolsep}{2.5pt}

\begin{tabular}{
>{\raggedright\arraybackslash}m{0.17\textwidth}
>{\raggedright\arraybackslash}m{0.26\textwidth}
>{\centering\arraybackslash}m{0.18\textwidth}
>{\raggedright\arraybackslash}m{0.34\textwidth}
}

\hline

\rowcolor{BlueBG}

\textbf{Category}
&
\textbf{Core Idea}
&
\textbf{Representative Methods}
&
\textbf{Key Features, Strengths, and Limitations}

\\

\hline


\rowcolor{BlueBG}

\raisebox{1pt}{\CategoryCell{
\textcolor{BlueText}{
\textbf{Single-Stream\\
Large-Patch\\
Transformer
}
}
}}

&
\LeftCell{

A single pathway directly processes large image patches.\\[1pt]

Large patches reduce visual token numbers,
while the Transformer directly predicts clean images.

}

&
\CenterCell{

\textbf{JiT}

}

&
\LeftCell{

\strength{
The architecture is highly streamlined and supports end-to-end training without requiring a VAE or separate decoder.
}

\vspace{2pt}

\limitation{
Large patches may discard fine-grained details, while small patches substantially increase computational and memory costs.
}

}

\\

\hline


\rowcolor{GreenBG}

\CategoryCell{
\textcolor{GreenText}{
\textbf{Hierarchical and\\
Hourglass pDiT
}
}
}

&
\LeftCell{

Multi-scale token organization first extracts local information at high resolution, aggregates global interactions at low resolution, and finally restores details through an upsampling pathway.

}

&
\CenterCell{

\textbf{HDiT}\\

}

&
\LeftCell{

\strength{
Global modeling is performed over shorter token sequences, improving efficiency while preserving global semantics and local details.
}

\vspace{2pt}

\limitation{
The architecture is more complex, and cross-level information transfer strongly affects performance.
}

}

\\

\hline


\rowcolor{OrangeBG}

\CategoryCell{
\textcolor{OrangeText}{
\textbf{Global--Local\\
Decoupled\\
Architecture
}
}
}

&
\LeftCell{

Large-patch DiT models global semantic layouts, while small-patch or lightweight modules recover local details, enabling explicit division of labor.

}

&
\CenterCell{

\textbf{DiP}\\
\textbf{PixelDiT}

}

&
\LeftCell{

\strength{
Alleviates the conflict between global modeling and high-fidelity detail reconstruction.
}

\vspace{2pt}

\limitation{
Usually requires additional detail modules or dual-level Transformers.
}

}

\\

\hline


\rowcolor{PurpleBG}

\CategoryCell{
\textcolor{PurpleText}{
\textbf{Frequency-Decoupled\\
Architecture
}
}
}

&
\LeftCell{

Low-frequency semantics and high-frequency details are modeled separately across different diffusion steps or frequency bands.

}

&
\CenterCell{

\textbf{DeCo}\\

}

&
\LeftCell{

\strength{
Explicit frequency separation improves global consistency and texture quality.
}

\vspace{2pt}

\limitation{
Frequency decomposition and optimization objectives require careful balancing.
}

}

\\

\hline


\rowcolor{CyanBG}

\CategoryCell{
\textcolor{CyanText}{
\textbf{Implicit Neural Field\\
Decoding Architecture
}
}
}

&
\LeftCell{

Large patches control token numbers, while implicit neural-field decoders reconstruct patch pixels in continuous coordinates.

}

&
\CenterCell{

\textbf{PixNerd}

}

&
\LeftCell{

\strength{
Represents rich details with fewer tokens and provides flexible continuous decoding.
}

\vspace{2pt}

\limitation{
Training and decoding costs may increase.
}

}

\\

\hline


\rowcolor{RedBG}

\raisebox{-7pt}{\CategoryCell{
\textcolor{RedText}{
\textbf{Cross-Scale Semantic\\
Anchoring Architecture
}
}
}}

&
\LeftCell{

Connects large-scale semantic tokens with small-scale detail tokens, allowing detail reconstruction to query high-level semantic features.

}

&
\CenterCell{

\textbf{HyperDiT}

}

&
\LeftCell{

\strength{
High-level semantics guide detail reconstruction and reduce structural inconsistency.
}

\vspace{2pt}

\limitation{
Cross-scale positional alignment remains challenging.
}

}

\\

\hline


\rowcolor{BlueBG}
\raisebox{-16pt}{\CategoryCell{
\textcolor{DeepBlueText}{
\textbf{Unified Multimodal\\
Shared-Token\\
Architecture
}
}
}}

&
\LeftCell{

Text tokens, image-condition tokens, and generation tokens are mapped into a shared space through a unified Transformer.
}
&
\raisebox{-8pt}{\CenterCell{\textbf{HiDream-O1-Image}
}}
&
\LeftCell{
\strength{
Shared representations improve generalization and data efficiency.\\
}
\vspace{2pt}
\limitation{
Task conflicts require careful attention allocation and capacity design.
}
}
\\
\hline
\end{tabular}
\caption{Taxonomy of modern pDiT architectures}
\label{tab:pdit_taxonomy}
\endgroup
\end{table*}

\begin{table*}[t]
\centering

{\Large
\bfseries
\textcolor{TitleBlue}{Core Challenges and Structured Solutions for pDiT}
}

\vspace{0.65em}

\begingroup
\renewcommand{\ChallengeCell}[1]{%
  \begin{minipage}[c]{\linewidth}
  \centering
  \setlength{\parindent}{0pt}%
  \vspace*{1pt}
  \noindent\ignorespaces#1\unskip\par
  \vspace*{1pt}
  \end{minipage}%
}
\renewcommand{\EssenceCell}[1]{%
  \begin{minipage}[c]{\linewidth}
  \raggedright
  \setlength{\parindent}{0pt}%
  \vspace*{2.5pt}
  \noindent\ignorespaces#1\unskip\par
  \vspace*{2.5pt}
  \end{minipage}%
}
\renewcommand{\SolutionCell}[1]{%
  \begin{minipage}[c]{\linewidth}
  \raggedright
  \setlength{\parindent}{0pt}%
  \vspace*{1pt}
  \noindent\ignorespaces#1\unskip\par
  \vspace*{1pt}
  \end{minipage}%
}
\renewcommand{\solutionitem}[2]{%
  \textcolor{SolutionText}{\bfseries #1: }#2%
}
\fontsize{8.3pt}{11.2pt}\selectfont
\setlength{\tabcolsep}{6pt}
\renewcommand{\arraystretch}{1.0}
\arrayrulecolor{LineBlue}

\begin{tabular}{
    >{\centering\arraybackslash}m{0.185\textwidth}
    >{\raggedright\arraybackslash}m{0.285\textwidth}
    >{\raggedright\arraybackslash}m{0.47\textwidth}
}

\rowcolor{ChallengeBlue}
\textcolor{white}{\bfseries Core Challenge}
&
\cellcolor{EssenceGreen}
\centering\arraybackslash
\textcolor{white}{\bfseries Underlying Issue}
&
\cellcolor{SolutionPurple}
\centering\arraybackslash
\textcolor{white}{\bfseries Structured Solution Pathways (Representative Strategies)}
\\
\hline

\rowcolor{ChallengeBG}
\ChallengeCell{
Quadratic Complexity of\\
High-Resolution Token Sequences
}
&
\EssenceCell{

High-resolution images produce long token sequences, causing the computational and memory costs of global self-attention to grow quadratically.
}
&
\SolutionCell{
\solutionitem{Reduce the Effective Sequence Length}
{Use large patches, semantic backbones, or hierarchical token aggregation to reduce the number of tokens involved in global attention.}






}
\\
\hline

\rowcolor{EssenceBG}
\ChallengeCell{
Granularity Conflict between\\
Semantic and Detail Scales
}
&
\EssenceCell{

Large patches yield shorter sequences, which benefit global semantic modeling but may lose local details.
Small patches preserve finer local structures but significantly increase the token count and computational cost.

}
&
\SolutionCell{
\solutionitem{Global--Local Decoupling}
{Use a large-patch DiT to model the global layout and lightweight detail modules to recover textures and edge information.}

\vspace{2pt}

\solutionitem{Dual-Level Transformer}
{Assign semantic structure modeling to a coarse-grained branch and local refinement to a fine-grained branch.}



}
\\
\hline

\rowcolor{ChallengeBG}
\raisebox{-12pt}{\ChallengeCell{
Difficulty Optimizing High-Dimensional Trajectories
}}
&
\EssenceCell{
\tablebullet
Directly regressing noise, velocity, or flow fields in the noisy state space produces a high-dimensional supervision target.

\vspace{2pt}



\tablebullet
Semantic structures form slowly, which can lead to difficult convergence or unstable optimization.
}
&
\SolutionCell{
\solutionitem{Direct Clean-Image Prediction}
{Adopt simpler and more direct learning targets, such as \(x_{0}\)-prediction.}

\vspace{2pt}

\solutionitem{Introduce Teacher Models}
{Use frozen vision foundation models to provide high-level semantic supervision and accelerate semantic structure formation.}




}
\\
\hline

\rowcolor{SolutionBG}
\raisebox{-14pt}{\ChallengeCell{
High-Frequency Signal and Noise Interference
}}
&
\EssenceCell{
\tablebullet
During early diffusion stages, high-frequency components are dominated by noise, while real image information remains weak.

\vspace{2pt}

\tablebullet
Uniformly modeling all frequency bands makes the model susceptible to interference from high-frequency noise.

}
&
\SolutionCell{

\solutionitem{Spectral Filtering and Frequency Scheduling}
{Dynamically select or adjust the learnable frequency range according to the diffusion timestep.}

\vspace{2pt}

\solutionitem{Frequency-Domain Loss Decomposition}
{Use low-frequency losses to enforce global consistency and high-frequency losses to preserve local details.}

}
\\
\hline

\rowcolor{EssenceBG}
\raisebox{-15pt}{\ChallengeCell{
Semantic Initialization and Representation Alignment
}}
&
\EssenceCell{
\tablebullet
Learning high-level semantic representations directly from random initialization is generally difficult.

\vspace{2pt}

\tablebullet
Frozen vision foundation models can provide stable semantic priors and a well-structured feature space.

}
&
\SolutionCell{
\solutionitem{Frozen VFM Guidance}
{Use visual features from clean images as teacher signals to guide intermediate representation learning.}

\vspace{2pt}



\solutionitem{Semantic Anchoring Rather Than Replacement}
{Use vision foundation models primarily for semantic initialization rather than replacing the pixel-space generation objective.}


}
\\
\hline

\end{tabular}

\vspace{0.65em}

\begin{tabular}{
    >{\centering\arraybackslash}m{0.095\textwidth}
    >{\raggedright\arraybackslash}m{0.865\textwidth}
}
\hline
\cellcolor{ChallengeBlue}
\textcolor{white}{\bfseries Overall Goal}
&
\cellcolor{GoalBG}
Balance computational scalability, training stability, global semantic consistency, and local-detail fidelity to build efficient, scalable, and general-purpose pixel-space diffusion Transformers.
\\
\hline
\end{tabular}

\endgroup

\caption{Core challenges and structured solutions for pDiT. To address computational efficiency, granularity conflicts, optimization difficulty, frequency interference, and semantic initialization, modern pDiTs employ structured designs to achieve scalable and high-fidelity pixel-space generation at high resolutions.}
\label{tab:pdit_challenges_solutions}

\end{table*}
Third, the scaling capability of Transformers provides a new backbone choice for pixel-space generation. Compared with convolutional U-Nets, Transformers can improve their modeling capacity by increasing model depth, hidden dimension, training data, and computational budget. Importantly, the attention mechanism can directly establish relations between distant patches, making it more suitable for expressing object relations, spatial layouts, and conditional dependencies in complex scenes. The rise of pDiTs can be understood as the joint result of three factors: stronger training recipes reduce the optimization difficulty of high-dimensional pixel space; more appropriate prediction parameterizations improve semantic learning and lower the overall optimization difficulty; and the scalability of Transformers enables models to learn global visual structures from large-scale data.

\subsection{Architectural Categories of Modern pDiTs}

From an architectural perspective, the central challenge of modern pDiTs is how to achieve global semantic modeling, faithful preservation of local details, and acceptable computational cost simultaneously, without relying on a fixed VAE. This survey categorizes existing methods according to how they model global structure, recover local details, and exchange information across scales. See Tab.~\ref{tab:pdit_taxonomy}.

\noindent \textbf{Single-Stream Large-Patch Transformer}:

The first category adopts a single-stream Transformer that directly processes raw pixel patches and controls the token count by using relatively large patch sizes. Its typical form is: 
\begin{equation}
H
=
\mathcal{T}_{\theta}
\left(
\mathcal{E}_{P}(\mathbf{x}_t),t,c
\right),
\quad
\hat{\mathbf{x}}_0
=
\mathcal{D}_{P}(H),
\end{equation}
where $\mathcal{T}_{\theta}$ denotes the Transformer backbone parameterized by $\theta$, while $\mathcal{E}_{P}$ and $\mathcal{D}_{P}$ denote the patch embedding module and the patch reconstruction head, respectively. JiT~\cite{li2026back} is a representative example of this paradigm. Its key idea is to use relatively large image patches to shorten the token sequence length and to alleviate the optimization difficulty in high-dimensional pixel space by directly predicting the clean image. 

The advantage of this paradigm lies in its structural simplicity. The model requires neither a VAE, nor cascaded super-resolution modules, nor an independent pixel decoder, yielding an end-to-end training path. Its limitations are evident: when the patch size is large, each token must carry more local texture and edge information, making it difficult for a simple linear output head to fully recover fine-grained structures; when the patch size is reduced, the number of tokens grows rapidly, making attention computation a new bottleneck. Therefore, single-stream large-patch pDiTs is better suited as a minimal architectural baseline, while providing a foundation for subsequent fine-grained decoding and hierarchical computation.

\noindent \textbf{Hierarchical and Hourglass pDiTs}:

The second category reduces the direct computational burden on high-resolution features through hierarchical token organization. The basic idea is to first extract local visual information at high-resolution levels, then progressively aggregate tokens into a lower-resolution semantic space for global interaction, and finally recover details through an upsampling path and cross-level connections. This process can be abstracted as:
\begin{equation}
H^{(s+1)}
=
\mathcal{M}^{(s)}
\left(
H^{(s)}
\right),
\quad
H^{(0)}
=
\mathcal{E}_{P}(\mathbf{x}_t),
\end{equation}
where \(\mathcal{M}^{(s)}\) denotes the token aggregation or downsampling module at the \(s\)-th stage. After Transformer computation at a lower resolution, the model propagates semantic features back to the high-resolution grid through a decoding path.

This class of methods inherits the multi-scale computation principle of U-Net, but replaces convolutional feature transformations with Transformer-based token interactions. HDiT is a representative architecture in this line. It organizes attention computation across different resolution levels through an Hourglass structure, allowing the model to avoid repeatedly performing expensive global interactions over the full high-resolution token grid. The core value of hierarchical design lies in decoupling high-resolution local processing from low-resolution global modeling, so that long-range semantic relationships are mainly established over shorter sequences, while detail recovery is achieved through cross-level connections.


\noindent \textbf{Global--Local Decoupled Architectures}:

The third category explicitly assigns global semantic modeling and local detail recovery to different modules. The motivation is that large patches help reduce the token count and establish global structure, but they sacrifice texture, edges, and small-scale object details; small patches are beneficial for high-fidelity recovery, but incur substantial attention overhead. Global--local decoupled architectures therefore use a large-patch DiT to model semantic layout, and then employ a lightweight detail module to compensate for local information:
\begin{equation}
H_{\mathrm{global}}
=
\mathcal{T}_{\mathrm{global}}
\left(
\mathcal{E}_{P_g}(\mathbf{x}_t)
\right),
\end{equation}
\begin{equation}
\hat{\mathbf{x}}_0
=
\mathcal{D}_{\mathrm{local}}
\left(
H_{\mathrm{global}},
\mathcal{E}_{P_l}(\mathbf{x}_t)
\right),
\quad
P_g > P_l.
\end{equation}
DiP uses a large-patch DiT to construct global structure and trains a Patch Detailer Head to recover local high-frequency details using context from the backbone. PixelDiT constructs a two-level architecture consisting of a patch-level DiT and a pixel-level DiT, assigning semantic layout and texture refinement to Transformer pathways at different scales. Such methods no longer require a single Transformer to undertake the generation of all frequency components and all spatial scales simultaneously. Instead, they alleviate the conflict between the semantic capability of large patches and the detail fidelity of small patches through a structured division of labor.

\noindent \textbf{Frequency-Decoupled Architectures}:

The fourth category reorganizes the pixel generation process from the perspective of frequency decomposition. In natural images, low-frequency components determine overall shape, layout, and large-scale color relationships, while high-frequency components carry texture, edges, and microscopic visual details. If use the same Transformer to model both types of signals throughout the diffusion trajectory, the model can spend substantial capacity on high-frequency noise and local errors, thereby weakening its learning of global semantics.

DeCo explicitly decouples low-frequency semantics from high-frequency textures, enabling the backbone DiT to focus on low-frequency structural modeling, while a lightweight pixel decoder restores high-frequency details under semantic guidance. Its training principle can be abstracted as:
\begin{equation}
\mathcal{L}
=
\lambda_{\mathrm{low}}
\mathcal{L}_{\mathrm{low}}
+
\lambda_{\mathrm{high}}
\mathcal{L}_{\mathrm{high}},
\end{equation}
where $\mathcal{L}_{\mathrm{low}}$ emphasizes semantic and structural recovery, while $\mathcal{L}_{\mathrm{high}}$ constrains texture and edge details. Spectral Forcing, by contrast, starts from the perspective of the input signal. According to the frequency distributions of the effective signal and noise at timesteps, it applies time-dependent spectral filtering to the noisy input, thereby preventing the model from paying excessive attention during early denoising stages to high-frequency regions that are mainly occupied by noise.
The significance of frequency-decoupled methods is that they no longer assume that every pixel error has equal learning value. Instead, by distinguishing global semantic signals, local texture signals, and high-noise frequency bands, the model assigns more appropriate generative responsibilities to different modules at different stages.

\noindent\textbf{Implicit Neural Field Decoding Architectures}:

The fifth category retains relatively short patch-token sequences, but no longer uses a simple linear layer to directly reconstruct the entire patch. Instead, each patch token is interpreted as the conditional representation of a local implicit neural field. For a relative coordinate \((\xi,\eta)\) within the \(i\)-th patch, the pixel value can be expressed as:
\begin{equation}
\hat{\mathbf{x}}_i(\xi,\eta)
=
f_{\theta}
\left(
\mathbf{z}_i,\gamma(\xi,\eta)
\right),
\end{equation}
where $\mathbf{z}_i$ denotes the conditional token corresponding to the $i$-th patch; $(\xi,\eta)$ denotes the two-dimensional relative coordinate within that patch; $\gamma(\cdot)$ denotes the coordinate encoding function; $f_{\theta}$ denotes the implicit neural field decoder conditioned on the local token and the coordinate encoding; and $\hat{\mathbf{x}}_i(\xi,\eta)$ denotes the predicted pixel value or color vector at that coordinate.
PixNerd~\cite{wang2025pixnerd} is a representative approach in this direction. It predicts local content in pixel space through patch-wise neural fields, attempting to avoid the detail loss caused by large-patch linear reconstruction. Compared with explicit pixel-level Transformers, implicit neural field decoding shifts part of the burden of detail modeling from sequence attention to a coordinate-conditioned decoding function, and thus has  computational potential for high-resolution generation.

\noindent\textbf{Cross-Scale Semantic Anchoring Architectures}:

The sixth category addresses the problem that fine-grained tokens lack stable global semantic constraints when recovering local pixels. Large-scale tokens are usually better at learning object categories, spatial relationships, and overall layout, whereas fine-scale tokens are needed to express edges, textures, and local geometry. Cross-scale semantic anchoring methods introduce explicit information pathways that allow fine-grained generation branches to query high-level semantic features.
HyperDiT~\cite{he2026hyperdit} connects large-scale semantic streams with small-scale detail streams through Hyper-Connected Cross-Scale Interaction, enabling fine-grained tokens to obtain global guidance from multi-level semantic anchors. Such methods must also address positional alignment across different patch scales, and therefore typically need to introduce scale-aware positional encodings or cross-scale geometric mappings. A frozen vision foundation model can provide a more stable semantic teacher signal during training, so that semantic anchors do not rely entirely on the noisy input itself. However, the compressed nature of external visual representations may also create information asymmetry with high-dimensional pixel recovery, and their use should be carefully designed according to the task and training stage.

\noindent\textbf{Shared-Token Unified Multimodal Modeling}:

The seventh category further extends pDiTs from a single-image generation backbone into a unified multimodal generative network. Its basic form is to map text tokens, conditional image tokens, and pixel tokens to be generated into a shared representation space.
These tokens are then jointly modeled by the same Transformer under the control of attention masks and task conditions. This design allows text-to-image generation, image editing, reference-image-driven generation, and visual understanding tasks to share the same context and backbone network. HiDream-O1-Image~\cite{cai2026hidream} is a representative example of this trend. It feeds text tokens, pixel-space generation tokens, and task-condition tokens into a unified Transformer, thereby supporting generation, editing, subject personalization, and related tasks within a shared architecture.

It should be noted that a unified token space does not require removing all modality-specific encoders. For example, reference-image conditions may still rely on a vision encoder to obtain semantic tokens, whereas the generation target itself is modeled directly in pixel space via diffusion. Therefore, the key to unified pDiTs is not the mechanical removal of all front-end modules, but rather avoiding a fixed VAE latent space as the upper bound that determines image-generation capability, and enabling supervision from different tasks to jointly shape visual representations within a shared backbone.

\subsection{Core Challenges and Structured Solutions for pDiTs}
Despite eliminating the compression bottleneck of latent diffusion models, pDiTs introduce new challenges arising from direct modeling in the high-dimensional pixel space. These challenges mainly involve computational complexity, multi-scale representation, optimization difficulty, frequency interference, and insufficient semantic initialization. See Tab.~\ref{tab:pdit_challenges_solutions}.

\noindent\textbf{Quadratic Cost of High-Resolution Token Sequences}:

Because pDiTs no longer rely on a pretrained visual compressor to reduce spatial resolution, they must directly model token sequences formed from high-resolution pixel patches. As the input resolution increases, the number of visual tokens grows rapidly, while standard global self-attention must model relationships between all token pairs, leading to substantial computational and memory overhead. This problem is exacerbated by the combined effects of high-resolution generation, deep Transformer stacks, and multi-step iterative sampling. Low-level optimizations such as FlashAttention~\cite{dao2022flashattention} can alleviate the storage pressure of intermediate attention matrices, but they do not fundamentally change the quadratic computational nature of global attention. Therefore, pDiT architectures must balance token granularity, global receptive field, and computational efficiency, and must avoid repeatedly applying global attention to the full high-resolution sequence through mechanisms such as large patches, hierarchical modeling, local attention, sparse interactions, or dynamic tokenization.

\noindent \textbf{Granularity Conflict Between Semantic and Detail Scales}:

High-resolution generation involves a granularity conflict. If large patches are used, the model can capture global semantic relationships with shorter sequences, but each token contains more local structure, making textures, text, and fine edges difficult to recover. If small patches are used, the model obtains finer-grained pixel control, but token count and attention cost increase sharply. This issue can be summarized as:
\begin{equation}
P \uparrow
\Rightarrow
N \downarrow
\Rightarrow
\text{global efficiency} \uparrow,
\end{equation}
\begin{equation}
P \downarrow
\Rightarrow
\text{local fidelity} \uparrow,
\quad
\mathcal{C}_{\mathrm{attn}} \uparrow.
\end{equation}
Global--local decoupled architectures, dual-level Transformers, implicit neural field decoding, and cross-scale semantic anchoring are all responses to this problem. Their common principle is to confine expensive global interactions to relatively coarse semantic tokens and then recover high-frequency details through local pathways, thereby avoiding the need to process all visual information at a uniform granularity.

\noindent \textbf{Optimization Difficulty of High-Dimensional Trajectories}:

Noisy states in pixel space span a high-dimensional continuous space, whereas real images usually concentrate near a relatively low-dimensional data manifold. If the model directly regresses noise or velocity, it must learn complex vector fields over many intermediate states far from the natural image manifold, which increases optimization difficulty and delays semantic convergence. Direct clean-image prediction, time-dependent loss reweighting, and perceptual supervision therefore become important solution paths.
Taking PixelGen as an example, the model augments direct image prediction with a local perceptual loss and a DINO-based global semantic loss, so that it no longer fits all pixel errors uniformly but instead prioritizes components that affect visual quality and semantic consistency. This line of thought indicates that the challenge of high-dimensional pixel space lies not only in variable count, but in whether the supervision signal can guide the model to prioritize image structures with perceptual significance.

\noindent \textbf{High-Frequency Signals and Noise Interference}:

In early diffusion stages, high-frequency components are often strongly contaminated by noise. If the model handles all frequency bands in the same way at all timesteps, it may expend a large computational budget on predicting low-value noise. Frequency-decoupled methods reduce this interference by allowing the backbone network to focus on low-frequency semantics while assigning high-frequency texture recovery to dedicated modules. Spectral filtering methods instead dynamically adjust the visible frequency range of the input according to the timestep, enabling the model to prioritize more reliable visual signals under the current noise level.
The key to this direction is not simply suppressing high frequencies, but reallocating frequency-modeling responsibilities according to the generation stage. Low-frequency structures usually determine object layout and overall semantics, whereas high-frequency structures determine texture clarity, edge sharpness, and text readability. An efficient pDiT should be able to process these two types of signals differently across timesteps and scales.

\noindent \textbf{Semantic Initialization and Representation Alignment}:

Direct pixel denoising loss can provide complete reconstruction supervision, but high-level semantics often require prolonged training before they become stable. To address this issue, some works introduce a frozen vision foundation model during training and pull the intermediate representations of the denoising network toward the semantic feature space of clean images. Let the frozen teacher network be \(F_{\omega}\), and let the intermediate features of the diffusion Transformer be \(\mathbf{H}_{\theta}^{(l)}\). The representation alignment objective can be written as
\begin{equation}
\mathcal{L}_{\mathrm{align}}
=
\frac{1}{N}
\sum_{i=1}^{N}
\left[
1-
\frac{
p_{\rho}(\mathbf{h}_i^{(l)})^{\top}
\mathbf{r}_i
}{
\left\|
p_{\rho}(\mathbf{h}_i^{(l)})
\right\|_2
\left\|
\mathbf{r}_i
\right\|_2
}
\right],
\end{equation}
where
\begin{equation}
\left[
\mathbf{r}_1,\ldots,\mathbf{r}_N
\right]
=
F_{\omega}(\mathbf{x}_0).
\end{equation}

This strategy can help the model establish object-level and structural semantics earlier. However, high-dimensional pixel reconstruction and compressed semantic representations are not fully consistent. If token-wise alignment is directly imposed throughout all training stages, the model may overly pursue low-dimensional teacher features and sacrifice sample diversity. Therefore, semantic alignment is more suitable as a stage-wise guide, a supervision signal for local modules, or a semantic anchoring mechanism, rather than as a replacement for the pixel-level generative objective itself.

\subsection{Section Summary}

Modern pDiTs should not be regarded as a simple reappearance of early convolutional pixel-space diffusion models, but should instead be understood as an independent and scalable class of generative models. It uses the raw pixel space as the diffusion domain, adopts Transformers as the primary modeling backbone, and addresses the systematic difficulties of high-dimensional generation through hierarchical token organization, global--local division of labor, frequency decoupling, implicit pixel decoding, cross-scale semantic interaction, and a shared multimodal token space.
From the perspective of technological evolution, the main problems of early pixel-space diffusion models were the computational cost of high-resolution modeling and insufficient semantic learnability. Latent diffusion alleviated the computational-efficiency problem to some extent through fixed VAE compression, but it also introduced an irreversible information bottleneck. Modern pDiTs instead seeks to eliminate dependence on fixed visual compressors and to achieve efficient, high-fidelity end-to-end generation through training objectives and architectural designs better suited to high-dimensional pixel space.



\section{Unified Multimodal Models using Pixel-Space Diffusion}
\label{sec:unified_pixel_diffusion}

Unified multimodal models aim to jointly process text, images, and task conditions through a shared network backbone, thereby supporting tasks such as image generation, instruction-based editing, subject personalization, and cross-modal contextual modeling. Existing methods typically rely on discrete visual tokenizers or pretrained VAEs to convert images into discrete visual tokens or continuous visual latents, which are then fed into a unified Transformer together with text tokens. Although such compressed representations can effectively shorten visual sequences and reduce generation cost, the visual generation process remains constrained by the quantization error of external tokenizers or the reconstruction ceiling of VAEs. Moreover, the unified backbone operates not on raw visual signals, but on intermediate representations predefined by an independent visual compressor.

Pixel-space diffusion provides an alternative technical pathway for unified multimodal modeling. Its core idea is not to treat each pixel as an individual token, but to partition the raw image into local patches and use learnable mappings to project pixel patches, text tokens, and task conditions into a shared hidden space. Unlike unified models that rely on fixed visual tokenizers, pixel-space models define the diffusion trajectory, prediction targets, and training supervision of the visual content to be generated directly in the raw image space, allowing visual representations and the unified Transformer to be optimized end-to-end within the same training process. Thus, the key shift in this paradigm is not simply the addition of another visual input format, but the removal of the fixed compression interface in the generation pathway, enabling the unified model to directly connect language semantics, task context, and pixel-level generation objectives.

\subsection{Native Pixel Tokens for Unified Multimodal Modeling}

For a noisy image $\mathbf{x}_t\in\mathbb{R}^{H\times W\times C}$, a unified pixel-space model first partitions it into non-overlapping $P\times P$ patches and obtains visual tokens through a learnable pixel mapper:
\begin{equation}
\mathbf{Z}_{\mathrm{pix}}
=
\mathcal{P}_{\omega}
\left(
\mathrm{Patchify}_{P}(\mathbf{x}_t)
\right),
\end{equation}
where $\mathrm{Patchify}_{P}(\cdot)$ denotes the operation that partitions the input image into local patches of size $P\times P$, $\mathcal{P}_{\omega}$ denotes a learnable pixel projection module parameterized by $\omega$, and $\mathbf{Z}_{\mathrm{pix}}$ denotes the sequence of pixel patch tokens constructed from the noisy image. Here, patchification is used only to organize the input sequence for the Transformer; it does not constitute an independently trained or fixed visual compressor.

Prompts, pixel patches, and task conditions are then mapped into a shared hidden space as a unified context sequence:
\begin{equation}
\mathbf{Z}
=
\left[
\mathbf{Z}_{\mathrm{text}};
\mathbf{Z}_{\mathrm{pix}};
\mathbf{Z}_{\mathrm{cond}}
\right],
\end{equation}
where $\mathbf{Z}_{\mathrm{text}}$ denotes the embedding sequence corresponding to discrete text tokens, $\mathbf{Z}_{\mathrm{pix}}$ denotes the pixel patch token sequence, and $\mathbf{Z}_{\mathrm{cond}}$ denotes the token sequence corresponding to reference images, editing instructions, subject information, or other task conditions. It is important to note that pixel-space unification does not require all conditions to enter the model in raw pixel form. Reference images and other complex conditions may still be converted into condition tokens through task-specific projection modules. The defining criterion of this paradigm is whether the diffusion state and supervision target of the visual content to be generated are defined directly in the raw pixel space, rather than whether all visual conditions adopt exactly the same encoding scheme.
The unified Transformer performs joint modeling conditioned on the diffusion timestep $t$, the shared context sequence, and task control signals:
\begin{equation}
\mathbf{H}
=
\mathcal{T}_{\theta}
\left(
\mathbf{Z},t,\tau
\right),
\end{equation}
where $\mathcal{T}_{\theta}$ denotes the unified Transformer backbone parameterized by $\theta$, $\mathbf{H}$ denotes its output joint hidden representations, $t$ denotes the diffusion or continuous-flow time variable, and $\tau$ denotes the control signal used to distinguish tasks such as image generation, image editing, and subject personalization. A pixel generation head then recovers the generation target from the corresponding visual positions:
\begin{equation}
\hat{\mathbf{x}}_0
=
\mathcal{D}_{\rho}
\left(
\mathbf{H}_{\mathrm{pix}}
\right),
\end{equation}
where $\mathbf{H}_{\mathrm{pix}}$ denotes the hidden states corresponding to the target pixel patches to be generated, $\mathcal{D}_{\rho}$ denotes the pixel reconstruction head parameterized by $\rho$, and $\hat{\mathbf{x}}_0$ denotes the predicted clean image. Depending on the specific training paradigm, the model may alternatively predict noise, velocity, or other parameterized targets equivalent to the pixel-space generation trajectory, but its final supervision is still applied directly in the raw image space.
HiDream-O1-Image~\cite{cai2026hidream} is a representative example of native pixel-space unified modeling. This model no longer relies on an external VAE or mutually separated pretrained text encoders. Instead, it directly maps raw image pixels, discrete text tokens, and task-specific conditions into a shared token space, and performs joint modeling through a Pixel-level Unified Transformer. Based on this unified representation, tasks such as text-to-image generation, instruction-based image editing, and subject-driven personalization can be organized as different forms of context-conditioned generation, without the need to construct a separate visual generation backbone for each task.

\subsection{Potential Value of Pixel Space for Unified Models}

As vision foundation models gradually move from single generation tasks toward unified paradigms of understanding and generation, a key question is how to construct shared representations that simultaneously support semantic reasoning and high-fidelity visual synthesis. Traditional multimodal models typically treat visual understanding and visual generation as distinct processes: understanding tasks rely on visual features with strong semantic abstraction, whereas generation tasks require precise spatial structure and fine-grained visual information. Therefore, maintaining semantic concepts, spatial relationships, and pixel-level details within the same model has become a central challenge for unified vision models.

Pixel-space diffusion provides a new representational paradigm for addressing this problem. Unlike methods that rely on predefined visual tokens or compressed latents, pixel-space models build the image generation process directly on raw visual signals, enabling the model to access both high-level semantic information and low-level visual evidence within a unified representation space. For unified models, this means that textual conditions, reference-image information, editing instructions, and generation targets can interact around a more complete visual representation, rather than being fused only after passing through different visual encoding pathways.

This unified representational capability is important for complex visual tasks. For example, in text-to-image generation, the model must not only understand object categories and semantic relationships, but also ground these semantic constraints in specific spatial locations, local shapes, and texture details. In image editing, the model must understand the intended modification while preserving structural consistency in unmodified regions. In reference-image-driven generation, the model must jointly model subject identity, visual style, and environmental relationships. These tasks all fundamentally require the model to access both abstract semantic information and fine-grained visual evidence, and pixel space provides a direct interface between the two.

Furthermore, pixel space promotes the transition of unified models from task-level fusion to representation-level fusion. In traditional architectures, understanding models obtain semantic features through a visual encoder, whereas generation models recover image content through an independent generative space, leaving an information transformation process between the two. Pixel-space unified models instead attempt to map text tokens, conditional visual information, and generative visual tokens into a shared Transformer context, so that language supervision, visual understanding objectives, and generation objectives jointly shape the same visual representation.

Therefore, the potential value of pixel space for unified models lies not only in improving generated image quality, but also in providing a shared representational foundation closer to the real visual world. Such a representation can simultaneously support visual understanding, image generation, editing control, and multimodal reasoning, offering a new pathway for building next-generation unified vision foundation models.

\subsection{Challenges Faced by Unified Models}

Native pixel-space modeling enables a unified Transformer to directly exploit visual information without the truncation imposed by fixed visual compressors, providing a higher upper bound on information fidelity for tasks such as text rendering, local editing, reference-based control, and fine-grained subject preservation. In parallel, raw pixel patches, text tokens, and task conditions can be jointly optimized end-to-end within a shared representation space, thereby reducing representational fragmentation among separate visual encoders, text encoders, and generation backbones, and providing a unified foundation for jointly modeling visual generation and contextual understanding. However, removing fixed visual compressors further amplifies challenges in high-dimensional pixel optimization, high-resolution computation, and multi-task co-training. The following discussion analyzes the core challenges faced by pixel-space unified models from these three perspectives.

\noindent \textbf{Optimization Difficulty in High-Dimensional Pixel Space:}

Latent space is not merely a mechanism for computational compression; it also provides a form of geometric prior. By mapping raw images into a lower-dimensional continuous latent space, a VAE often compresses high-frequency variations, local noise, and certain hard-to-reconstruct details into smoother representations. Although this process may lose visual information, it also reduces the spatial complexity that subsequent generative models need to model.
In contrast, pixel-space unified models must directly handle the noised high-dimensional pixel-space variable:
\begin{equation}
\mathbf{x}_t
=
\alpha_t \mathbf{x}_0
+
\sigma_t \epsilon,
\end{equation}
where \(\mathbf{x}_t \in \mathbb{R}^{H \times W \times C}\) denotes a high-dimensional noisy image. The model must recover not only object-level semantics and spatial layout, but also pixel-scale color fluctuations, texture changes, edge details, and noise perturbations. Some high-frequency variations lack stable semantic value for understanding tasks, yet cannot be completely ignored for high-fidelity generation. Therefore, a unified Transformer must simultaneously learn semantic invariance and pixel-level sensitivity within the same network capacity.

This issue can be summarized as a conflict between the high-dimensional pixel manifold and compressed priors. Latent-space models obtain a more compact and easier-to-optimize visual domain through a VAE, but at the cost of losing some details. Pixel-space models preserve more complete visual evidence, but must learn the denoising vector field over a more complex distribution. For large-scale unified models, this creates a significant optimization burden: if model capacity is insufficient, the visual generation branch may allocate a large number of parameters to fitting subtle pixel variations, thereby delaying the formation of language semantics, object relationships, and cross-modal alignment capabilities.

Therefore, modern pixel-space unified models generally cannot rely solely on raw pixel-level losses. A more reasonable approach is to introduce multi-scale computation, frequency-based division of labor, perceptual supervision, or semantic alignment, letting the model establish stable global semantic structures during early training and then progressively recover local pixel details. The core objective is to dynamically allocate the model's responsibility for modeling visual information at different scales throughout end-to-end training.


\noindent \textbf{Compute and Memory Costs of High-Resolution Modeling:}

Pixel-space unified models overcome the information limitations of fixed visual tokenizers, but they also introduce new computational scaling problems. The fundamental reason is that the model no longer learns only compressed visual semantics; instead, it must directly maintain structural information, local details, and cross-modal relationships in a high-dimensional visual space.
For a pixel-space Transformer, the number of image tokens satisfies:
\begin{equation}
N_{\mathrm{pixel}}
=
\frac{HW}{P^2}.
\end{equation}
As resolution increases, visual token count grows rapidly, and global attention must compute all-token relationships:
\begin{equation}
\mathcal{C}
=
O(N_{\mathrm{pixel}}^2d).
\end{equation}

Compared with pure image generation models, unified models further increase the computational dimensionality. Because text understanding, visual generation, and conditional control share the same Transformer, the model in practice processes a fused multimodal sequence. Therefore, high-resolution pixel-space models face not only an increase in the number of visual tokens, but also an increase in overall computational complexity caused by the expansion of cross-modal context. This challenge requires future pixel-space unified models to advance from ``unified representation'' toward ``unified and efficient computation mechanisms.'' On the one hand, models need to exploit hierarchical structures to reduce ineffective computation, allowing global semantic modeling and local detail recovery to be performed at different scales. On the other hand, they need to dynamically adjust token density according to the importance of information during generation, so that limited computational resources are prioritized for regions with high semantic value and high visual sensitivity.

\noindent \textbf{Multi-Task Gradient Interference in Unified Models:}

Unified models face gradient interference among different task objectives. Language understanding typically relies on cross-entropy loss over discrete tokens, whereas image generation typically relies on diffusion losses or flow-matching losses over continuous variables. Their objective functions, gradient scales, and optimization geometries all differ substantially. The unified training objective is commonly written as:
\begin{equation}
\mathcal{L}_{\mathrm{UMM}}
=
\lambda_{\mathrm{text}}
\mathcal{L}_{\mathrm{text}}
+
\lambda_{\mathrm{vision}}
\mathcal{L}_{\mathrm{vision}}
+
\lambda_{\mathrm{align}}
\mathcal{L}_{\mathrm{align}},
\end{equation}
where \(\mathcal{L}_{\mathrm{align}}\) can represent image--text alignment, semantic distillation, or task consistency constraints. Let \(\theta_{\mathrm{shared}}\) denote the shared backbone parameters. The gradients of the text and vision tasks in this parameter space are respectively:
\begin{equation}
g_{\mathrm{text}}
=
\nabla_{\theta_{\mathrm{shared}}}
\mathcal{L}_{\mathrm{text}},
\quad
g_{\mathrm{vision}}
=
\nabla_{\theta_{\mathrm{shared}}}
\mathcal{L}_{\mathrm{vision}}.
\end{equation}
The local optimization relationship between the two tasks can be measured by cosine similarity:
\begin{equation}
\rho
=
\frac{
g_{\mathrm{text}}^{\top}
g_{\mathrm{vision}}
}{
\left\|
g_{\mathrm{text}}
\right\|_2
\left\|
g_{\mathrm{vision}}
\right\|_2
}.
\end{equation}
When \(\rho < 0\), the two tasks exhibit direct gradient conflict in the current parameter region. The language branch may encourage the shared attention layers to strengthen abstract symbolic relations and causal dependencies, whereas the pixel-generation branch may require the same layers to preserve spatial locality, temporal conditioning, and high-frequency visual variations. If visual generation is performed in the original pixel space, this conflict is further intensified because the visual loss covers more tokens, and each token corresponds to denser continuous information.
Therefore, pixel-space unified models cannot rely only on a simple summation of losses. Their training usually requires a clearer division of responsibilities among tasks. For example, text and images may use different output heads, language tokens may adopt causal attention, and image tokens may use bidirectional attention within local image patches. Gradient competition can also be mitigated through staged training, dynamic loss weighting, modality-specific projection layers, and partial parameter isolation. Transfusion handles discrete text and continuous visual latents through modality-specific encoder--decoder layers and hybrid attention masks; Show-o2 uses separate language and flow-matching heads, and gradually learns generation capabilities and unified multimodal capabilities through two-stage training. These designs indicate that the goal of unified models is not to make all modalities fully share every path, but rather to preserve necessary structural differences for different modalities while maintaining a shared reasoning backbone.

\subsection{Why Pixel-Level Unified Models}

Although pixel-space unified models face greater optimization difficulty, higher computational costs, and stronger multi-task gradient interference, recent research remains interested in this direction because it may offer three potential benefits.

First, pixel space can reduce the constraints imposed by fixed VAEs or discrete codebooks on generation quality. For tasks such as complex text rendering, local editing, object identity preservation, and fine-grained geometric control, if visual information is compressed or quantized during encoding, it is difficult for the model to fully recover it. Pixel-level generation allows the model to receive supervision directly from pixel losses and perceptual losses, thereby preserving a larger expressive space for high-fidelity visual generation.

Second, pixel space provides a more direct condition for sharing visual evidence between understanding and generation. Traditional unified models often need to establish additional interfaces between abstract understanding features and high-fidelity generative latents, whereas pixel-level tokens allow local details, spatial structures, and semantic relationships to be jointly modeled within the same generative backbone. This is particularly important for OCR~\cite{wei2024general}, image editing~\cite{xiao2025omnigen}, reference-based control~\cite{wang2024instantid}, and multi-image contextual reasoning~\cite{jiang2024mantis}, because these tasks simultaneously depend on abstract semantic judgment and fine-grained visual fidelity.

Finally, structured computation in modern pDiT architectures partially mitigates the historical problems of raw pixel space. A large-patch semantic backbone can establish global layout over a shorter sequence, local refinement modules can recover high-frequency textures, cross-scale attention can semantically anchor fine-grained tokens, and frequency-aware training can reduce ineffective computation in noise-dominated regions. Thus, pixel-level unified models do not attempt to simply replace VAEs with greater computation. Instead, they seek to control the computational burden of high-dimensional modeling through multi-scale division of labor and end-to-end joint optimization while preserving pixel fidelity.

Overall, the current research trend is not to replace all unified models with pDiTs, but to explore new balances among discrete visual tokens, continuous latent representations, and pixel-level diffusion. Show-o demonstrates a unified formulation of discrete diffusion and language modeling; Transfusion and Show-o2 demonstrate hybrid generation paradigms over continuous latents; and HiDream-O1-Image indicates that a shared pixel-level token space is becoming a feasible new direction. The key question for the future is not whether to completely eliminate all visual encoders, but how to achieve efficient multi-scale token organization, stable cross-modal alignment, reasonable task-gradient coordination, and scalable high-resolution visual generation within a shared Transformer.

\section{pDiTs Applications}

Pixel-space diffusion models have continued to improve in training stability, architectural design, and computational efficiency, and their application scope has gradually expanded from standard image generation to high-fidelity visual tasks such as image editing, video generation, 3D content modeling, and medical imaging. Compared with latent diffusion, pDiTs does not rely on a fixed VAE to compress targets into low-dimensional latents. Instead, it defines the generative process directly over raw pixels, voxels, point maps, or pixel-aligned continuous attributes. Therefore, when evaluation criteria depend heavily on local textures, precise edges, spatial correspondences, or subtle structures, pixel-space modeling offers a more direct advantage. At the same time, ``pixel space'' in different applications does not always refer to two-dimensional RGB images. It may also denote voxel-level medical data, per-pixel 3D point maps, or 3D Gaussian attributes organized on a two-dimensional grid. Their shared characteristic is that the diffusion target operates directly on high-fidelity, task-relevant raw representations rather than on compressed latents constructed by a pretrained autoencoder.

\subsection{High-Fidelity Image Generation}

The earliest applications of pDiTs focused on class-conditional image generation, primarily to verify whether models could directly recover natural images from noise without a VAE. Simple Diffusion~\cite{hoogeboom2023simple}, SiD2~\cite{hoogeboom2025simpler}, HDiT~\cite{crowson2024scalable}, and JiT~\cite{li2026back} have demonstrated competitive generation results on benchmarks such as ImageNet, showing that end-to-end pixel diffusion can effectively generate two-dimensional images with clear subjects, coherent layouts, and natural textures. Subsequent works, including PixelDiT~\cite{yu2026pixeldit}, DeCo~\cite{ma2026deco}, PixNerd~\cite{wang2025pixnerd}, and HyperDiT~\cite{he2026hyperdit}, further improve global structure and local details in complex scenes, gradually enabling pixel-space models to support higher-resolution image generation.

As model scale and training data expand, the application scope of pDiTs has begun to shift from closed-set class-conditional generation to open-domain text-to-image synthesis. In this setting, the model must determine the image subject, spatial layout, color style, and local textures from natural-language prompts, while also handling multi-object composition, attribute binding, and complex relational expressions. Direct pixel generation can preserve fine-grained visual information that is easily lost in compressed representations, and therefore has potential advantages in high-fidelity scenarios such as text rendering, sharp edges, repetitive textures, and small-scale object generation. Thus, high-fidelity image generation is not only the earliest validated application of pDiTs, but also a primary foundation for its development toward general-purpose visual generative models.

\subsection{Image Editing, Inpainting, and On-Device Applications}

Image editing and restoration constitute another class of tasks for which pixel-space generation has direct practical potential. Unlike generating a complete image from noise, editing requires the model to satisfy two objectives simultaneously: on the one hand, it must modify specified regions according to text, masks, or reference images; on the other hand, it must strictly preserve the color, texture, identity, and geometric structure of unedited regions. Any encoding--decoding error may propagate to regions that should not be modified. Therefore, establishing conditional generative mappings directly in pixel space provides a more natural way to constrain local correspondences between input and output. Early work such as Palette~\cite{saharia2022palette} has shown that a unified pixel-space image-to-image diffusion model can support multiple tasks, including inpainting, colorization, deblurring, and JPEG restoration. Modern pDiT architectures are expected to further leverage Transformers' long-range modeling to combine local editing constraints with global semantic consistency.

Another important application for image editing is local inference on edge devices. Personal photos often contain sensitive data including identity, location, and living environment, and uploading them to servers may introduce privacy risks~\cite{shen2025generative}. BlazeEdit~\cite{deng2026blazeedit} integrates object removal, outpainting, tone correction, relighting, and sticker generation into a lightweight image-to-image diffusion model, enabling low-latency mobile inference. This suggests reducing text encoders, compressing model scale, and adopting few-step sampling can substantially improve the practicality of local editing.

It should be noted that BlazeEdit itself still follows the latent diffusion paradigm rather than pDiTs. However, its relevance for pDiTs lies in revealing the core requirements of edge-side image editing: privacy, low latency, small model size, and fidelity in non-edited regions. Future pDiT models for mobile devices need to preserve raw pixel details while further incorporating structured token compression, local sparse attention, quantization, and few-step distillation. Otherwise, the memory and computational costs introduced by high-resolution pixel sequences will continue to limit practical deployment.

\subsection{Video Generation and Spatiotemporal Modeling}

Video is a natural extension of pixel-space diffusion, but it introduces a more severe sequence-length problem. If a video contains \(T\) frames, each with resolution \(H\times W\), and the spatial patch size is \(P\times P\), visual token count is approximately:

\begin{equation}
N_{\mathrm{video}}
=
T\frac{HW}{P^2}.
\end{equation}
Under standard global self-attention, computational complexity grows rapidly with spatial resolution and temporal length. Therefore, video pDiTs must not only recover high-frequency details per frame, but ensure temporal consistency in motion trajectories, subject identity, and background structure.

SiD2~\cite{hoogeboom2025simpler} extends end-to-end pixel-space diffusion to Kinetics-600 video generation, showing that, with appropriate loss weighting and network scaling, pixel-space methods are not limited to static images. However, compared with latent-space video diffusion, native pixel-space video generation remains at a relatively early stage. Future methods may need space--time factorized attention, temporal hierarchical tokens, keyframe guidance, and local motion modeling, further extending global--local decoupling in static-image pDiTs into a three-way division of labor across semantics, space, and time.

\subsection{3D Generation and Reconstruction}

Three-dimensional content modeling has recently become one of the fastest-growing extensions of the pixel-space paradigm. Traditional 3D generation methods typically use latent diffusion to generate multi-view images or compressed 3D representations, and then obtain the final geometry through a separate reconstruction module. This multi-stage design can easily accumulate multi-view inconsistencies, latent decoding errors, and geometric reconstruction errors. Pixel-space methods instead attempt to perform denoising directly on high-fidelity representations corresponding to the final 3D structure, allowing appearance and geometric supervision to act more directly on the generative trajectory.

PixGS~\cite{cao2026pixgs} applies pixel-space diffusion to directly generate 3D Gaussian Splats. Rather than relying on latent models to indirectly produce 3D content, it directly denoises Gaussian attributes organized on a two-dimensional grid, using normal, depth, and high-frequency structural supervision to constrain appearance and geometry. Because each grid location corresponds to specific Gaussian attributes, the model can impose splat-level structural regularization during sampling, thereby reducing error accumulation from multi-stage generation.

PixWorld~\cite{gao2026pixworld} further unifies 3D scene generation and reconstruction within a single pixel-space diffusion framework. Its core idea is to define diffusion supervision directly on rendered images and to connect 2D observations with 3D scenes through pixel-aligned 3D Gaussian representations. Compared with defining a unified objective on latent features, this design enables reconstruction and generation to share supervision signals oriented toward final scene fidelity. PixWorld also introduces geometry-aware losses using 3D foundation models to compensate for the insufficient constraints that purely 2D photometric losses impose on real geometric structure.

For single-image 3D reconstruction, PointDiT~\cite{xu2026pointdit} applies the diffusion process directly to raw 3D point-map patches and uses pretrained DINOv3 image features as conditions. This method neither compresses the target point map nor requires an additional geometric tokenizer. Instead, it uses a standard ViT to directly predict geometric results on pixel-aligned 3D coordinate maps. Its significance lies in showing that the ``raw space'' of pDiTs can extend from RGB pixels to per-pixel geometric variables. In particular, as long as the target representation has a regular spatial organization, direct diffusion may avoid boundary blurring and flying-pixel artifacts caused by latent decoding.

Overall, PixGS, PixWorld, and PointDiT respectively represent three application pathways: direct 3D attribute generation, unified generation--reconstruction, and monocular geometric estimation. Together, they reflect a trend in which pixel-space diffusion is evolving from a two-dimensional image generator into a general modeling tool that connects image evidence with continuous 3D world representations.

\subsection{Volumetric Medical Image Generation and Translation}

In medical image generation, pixel-level fidelity has far greater clinical significance than in general natural-image generation. Minor texture deviations in natural images may affect only subjective visual quality, whereas subtle nodules, tissue boundaries, vascular structures, or density variations in medical images may directly correspond to pathological information. Therefore, medical generative models must satisfy not only visual realism, but also anatomical plausibility and pathological plausibility. Small structural losses caused by fixed autoencoders may have more serious consequences in such tasks, providing an important motivation for direct modeling in pixel or voxel space.

However, high-resolution 3D medical imaging also constitutes one of the most challenging application scenarios for pixel-space diffusion. Increasing the 3D resolution from \(128^3\) to \(256^3\) increases the number of voxels by a factor of eight. If global attention is applied directly to all voxel tokens, the computational and memory costs further rise sharply. Early work such as MedSyn~\cite{xu2024medsyn} adopts a cascaded diffusion strategy to generate high-fidelity 3D CT volumes: it first generates lower-resolution 3D volumetric data in pixel space, and then progressively restores high-resolution details using a separate pixel-space super-resolution diffusion model. This design avoids completing the entire generation process from pure noise at the highest resolution, but it also introduces multi-stage training and error propagation issues.

PRDiT~\cite{zhang2026pixel} directly targets voxel-level 3D CT generation through a coarse-to-fine two-stage architecture composed of local denoisers and a global residual Diffusion Transformer. The local branch efficiently estimates low-frequency structures on overlapping 3D patches, while the global Transformer uses memory-efficient attention to model high-frequency residuals across the entire volume. By explicitly separating local base structures from global high-frequency corrections, this method avoids the autoencoder bottleneck while reducing the optimization difficulty of high-resolution 3D generation.

PPDM~\cite{chen2026ppdm} targets high-resolution volumetric medical image translation and proposes a reversible pixel puzzle--unpuzzle operation, which rearranges part of the spatial resolution into the channel dimension, thereby reducing activation memory without using lossy latent compression. Its direct bridge diffusion starts modeling from a conditional medical image rather than from pure noise, so that the network mainly learns residuals related to the target modality. The puzzle-gradient loss is used to suppress grid artifacts that may be caused by spatial rearrangement. PPDM shows that pixel space does not necessarily mean that all computation must be performed on the original grid; reversible and information-preserving spatial rearrangements can be used to improve the speed and memory efficiency of diffusion for volumetric data.

These works show that the core problem of medical pDiTs is not simply to pursue better natural-image metrics, but to preserve diagnostically relevant visual evidence under computationally affordable conditions. Future evaluation protocols should jointly report 3D generation quality, anatomical consistency, lesion preservation rate, downstream diagnostic performance, and generation uncertainty, rather than relying only on 2D FID or slice-level visual assessment.

\subsection{Section Summary}

Existing applications indicate that the advantages of pDiTs are primarily concentrated in tasks that are highly sensitive to information loss. Two-dimensional image generation focuses on texture, text, and edges; image editing requires strict preservation of non-edited regions; video generation needs to maintain cross-frame consistency of fine details; 3D generation and reconstruction rely on accurate appearance--geometry correspondence; and medical imaging requires preserving subtle visual evidence related to anatomy and pathology. Although different applications adopt different data representations and network structures, they all attempt to avoid having a fixed compressor predetermine the upper bound of information by directly modeling the task-relevant raw space.

At the same time, the expansion of applications has further exposed the computational bottlenecks of pDiTs. As data extend from two-dimensional images to videos, 3D scenes, and volumetric medical data, the number of tokens and the cost of their interactions increase rapidly. Therefore, future application research should not only demonstrate that the ``raw space can be modeled,'' but also construct task-adaptive mechanisms for reversible tokenization, cross-scale interaction, sparse attention, local--global division of labor, and dynamic computation. Only when the gains in pixel-level fidelity outweigh the additional training and inference costs can pDiTs establish a stable advantage over latent-space models in practical applications.

\section{Discussion}

This section discusses key issues in the future development of modern visual generative models from three interrelated perspectives. First, pixel-space diffusion Transformers (pDiTs) reintroduce end-to-end pixel modeling capability, but at a substantial computational cost. Second, as vision foundation models increasingly serve as an important source for improving generation quality, a new challenge arises: how to exploit external semantic and structural priors without disrupting the representation space of the generative model itself. Finally, as visual generative models further evolve into unified models for understanding and generation, coordinating different task objectives within a shared Transformer while preventing the forgetting of original generative capabilities becomes key to building next-generation vision foundation models. Accordingly, this section discusses the computation--fidelity trade-off, non-invasive structural optimization, and capability preservation driven by joint objectives. It analyzes the strengths and limitations of current approaches and further outlines possible development paths for future unified vision models.

\subsection{The Latent--Pixel Compute--Fidelity Trade-off}

Although pixel-space diffusion Transformers overcome the representational constraints introduced by latent-space compression and exhibit stronger potential for end-to-end modeling, their advantages do not come without cost. Compared with latent diffusion models that rely on VAE or VQ tokenizers, pDiTs must directly process higher-dimensional visual token sequences, thereby introducing greater computational overhead during both training and inference. This issue is particularly pronounced in high-resolution generation, where the number of pixel-space tokens grows rapidly with image size, imposing heavier attention computation and memory pressure on the Transformer backbone. A key question is therefore whether the additional computational cost of pDiTs is truly converted into sufficiently significant gains in visual quality.

Regarding computational efficiency, latent-space diffusion retains clear advantages. Using a pretrained visual tokenizer, LDM maps raw images into more compact latent representations, enabling the diffusion model to perform generation in a lower-dimensional space. This design effectively reduces training cost and sampling overhead, and has become the dominant solution for large-scale visual generation systems. However, latent-space compression also limits the range of visual information accessible to the model. Because the generation process must ultimately pass through a latent-to-pixel decoder, the model's ability to recover fine-grained structures is constrained by the representational capacity of the tokenizer.

In contrast, pDiT allocates computation directly to modeling original visual space, allowing the model to bypass the information bottleneck from a fixed encoder. This design shows potential advantages in demanding visual tasks, including complex text rendering, fine-grained background consistency preservation, and local texture recovery. In these scenarios, generation quality depends not only on high-level semantic correctness, but also on whether the model can preserve complete spatial layout relationships and pixel-level visual evidence. Thus, the value of pDiT does not lie in comprehensively replacing latent-space models, but in exchanging higher computational cost for a higher upper bound on visual fidelity.

From an optimization perspective, latent-space models and pixel-space models represent two different directions. Latent-space models emphasize computational efficiency, whereas pDiT places greater emphasis on the upper bound of generation quality. Future model selection should not rely on a single metric, such as FID or inference speed, but should instead comprehensively consider training FLOPs, inference cost, and task-relevant quality metrics. For general image generation tasks, latent-space models may still retain stronger efficiency advantages. However, for tasks such as text rendering, fine-grained editing, reference preservation, and high-fidelity visual reconstruction, pDiT may obtain more competitive quality gains through stronger detail-modeling capability.

Therefore, the future development of pixel-space diffusion should not be directed toward unbounded increases in computational scale. Instead, while preserving the advantages of pixel-level information, it should reduce additional costs through hierarchical token design, dynamic computation allocation, local attention, and multi-scale generation strategies. A truly competitive pDiT should establish a new balance between computational efficiency and visual fidelity.

\subsection{Representation Tuning vs. Noninvasive Structure Guidance}

The preceding sections show that vision foundation models can improve diffusion training, but their representations should not be considered interchangeable with those required for generation. The central design choice is whether semantic priors should reshape the visual tokenizer or guide the generator without redefining its internal feature space. Tokenizer refinement, including VFM-based autoencoders, can improve semantic organization and reduce latent-diffusion optimization burden. However, it remains bounded by the information retained at the compression interface: improving semantic structure does not guarantee recovery of text, local geometry, or high-frequency appearance discarded during encoding.

Non-invasive structural guidance serves a complementary role. Rather than enforcing token-wise feature imitation, it should transfer only task-relevant relational information, such as object layout, region correspondence, and spatial consistency. This principle respects the distinct objectives of discriminative and generative representations: the former favors invariance and semantic abstraction, whereas the latter must preserve controllable visual variation. For pDiT, structural priors are therefore most useful as auxiliary constraints during global structure formation, rather than as a substitute for native pixel-space generative supervision.
A promising direction is to make this guidance adaptive rather than fixed. Its source, strength, and spatial scope should vary with diffusion time, image scale, and task requirements: stronger during semantic layout formation and weaker during high-frequency detail restoration. Evaluation should separately measure structural correctness, text fidelity, local-edit preservation, diversity, and computational overhead. The central question is thus not whether VFM features or pixel-space representations are superior, but how semantic priors can improve global organization without becoming a bottleneck for generative detail and diversity.

\subsection{Joint-Objective Forgetting Mitigation vs. Weight Cloning}

As visual generative models increasingly evolve toward unified understanding and generation, a key challenge is how to introduce new task capabilities while preserving the original generative ability. Unlike traditional single-task diffusion models, unified models must simultaneously optimize multiple objectives, such as language understanding, image generation, image editing, and conditional control, while different tasks often differ in their representation spaces and optimization directions. If a model adapts to new multimodal tasks only through simple fine-tuning, the original generative distribution can easily shift, leading to degraded generation quality, loss of fine details, and weakened conditional responsiveness.

One intuitive solution is Weight Cloning, which copies the parameters of an existing generative model into the unified model as initialization so that the new model inherits the original generative capability. This strategy can be written as
\begin{equation}
\theta_{\mathrm{new}}^{0}
=
\theta_{\mathrm{gen}},
\end{equation}
where \(\theta_{\mathrm{gen}}\) denotes the parameters of the pretrained generative model, and \(\theta_{\mathrm{new}}^{0}\) denotes the initialized parameters of the unified model.
However, Weight Cloning essentially provides only parameter initialization and cannot guarantee that generative capability will be continuously preserved during training. As multimodal objectives continue to update the model, the shared parameters are still affected by gradients from new tasks:
\begin{equation}
\theta_{\mathrm{new}}
=
\theta_{\mathrm{new}}^{0}
-
\eta
\nabla_{\theta}
\mathcal{L}_{\mathrm{multi}},
\end{equation}
where \(\mathcal{L}_{\mathrm{multi}}\) denotes the joint multitask objective. When understanding tasks or newly introduced conditional tasks dominate the optimization direction, the model parameters may gradually deviate from the original generative distribution, producing a problem analogous to catastrophic forgetting.

In contrast, methods based on Joint Objectives place greater emphasis on the functional constraints among different capabilities. The core idea is to retain both the generative objective and the newly introduced task objectives during unified training, enabling the model to learn new capabilities while maintaining its original visual generation ability. The unified optimization objective can typically be expressed as
\begin{equation}
\mathcal{L}_{\mathrm{joint}}
=
\lambda_{\mathrm{gen}}
\mathcal{L}_{\mathrm{gen}}
+
\lambda_{\mathrm{text}}
\mathcal{L}_{\mathrm{text}}
+
\lambda_{\mathrm{edit}}
\mathcal{L}_{\mathrm{edit}}
+
\lambda_{\mathrm{align}}
\mathcal{L}_{\mathrm{align}}.
\end{equation}
The generation loss preserves the capacity for image-distribution modeling, the text-understanding loss learns linguistic semantics, and the editing and alignment losses enhance cross-modal interaction.

Furthermore, joint objectives not only constrain final output quality but also help form more stable visual representations within the shared Transformer. Compared with directly copying parameters, Joint Objectives use continuous functional supervision to constrain the model to preserve its original generative trajectory while gradually absorbing new semantic capabilities. Therefore, future development of unified visual models should not simply expand the proportion of shared parameters; rather, it should design better task-coordination mechanisms within a shared representation space, allowing generation, understanding, and conditional control capabilities to coexist stably in a single model over the long term.

\subsection{Summary of the Discussion}
There is no task-independent absolute superiority between latent-space diffusion and pDiT. LDMs retain clear advantages in large-scale training and efficient deployment, whereas the value of pDiT mainly arises in scenarios where a fixed tokenizer has become a bottleneck for detail fidelity, text rendering, or end-to-end unified modeling. For pDiT, its high computational cost is sufficiently justified only when it demonstrates a clear empirical Pareto advantage in dimensions such as text quality, background consistency, and texture fidelity.

At the same time, priors from vision foundation models should be introduced without disrupting the native representational geometry of generative models. Compared with hard token-wise feature matching, SGA provides weaker and more compatible structural constraints through spatial self-similarity relationships. Latent optimization methods such as VA-VAE can improve the learnability of compressed representations, but they cannot fundamentally remove the information ceiling imposed by the compression interface. Finally, to avoid forgetting, unified multimodal models should not rely solely on cloning pretrained weights. Instead, they should jointly optimize generation objectives, understanding objectives, structural constraints, and functionality-preservation terms, thereby continuously coordinating visual fidelity and cross-modal capabilities within a shared Transformer.

\section{Future Outlook \& Open Challenges}

This survey does not argue that Pixel-Space Diffusion Transformers (pDiT) will completely replace latent diffusion models. Rather, it suggests that the main bottleneck in visual generation is shifting from generative capability alone toward a comprehensive trade-off among information fidelity, computational efficiency, and unified modeling. Latent diffusion obtains high computational efficiency through a fixed VAE, but in doing so fixes part of the upper bound on visual quality within an external compression interface. pDiT removes this interface, but must directly confront high-dimensional pixel modeling, high-resolution computation, and conflicts in multi-task optimization. The key question for future research is therefore not a simple choice between ``latent'' and ``pixel'', but the design of new visual generative architectures that can dynamically coordinate semantic abstraction, pixel fidelity, and computational budget according to task requirements.

\subsection{Dynamic Token Granularity and Adaptive Routing}

Existing pDiT models typically adopt fixed patch sizes and relatively static computation paths, making them poorly suited to the changing information demands of the diffusion process. In the early stages of generation, the model mainly needs to establish the global layout, object relationships, and low-frequency structure; in the later stages, the computational focus gradually shifts toward text strokes, object boundaries, and local textures. Future models should dynamically adjust token granularity according to the timestep, regional uncertainty, and semantic importance, so that limited computational resources are preferentially allocated to regions that remain unstable or have higher visual value.
The main challenge in this direction is maintaining consistency across scales during generation. Token splitting, merging, or dynamic routing may introduce boundary drift, local jumps, and texture discontinuities. Future methods therefore need more stable mechanisms for cross-scale state propagation and positional correspondence. The goal of dynamic computation should not merely be to reduce the number of tokens, but to enable adaptive allocation of computation from global composition to local refinement without disrupting the continuity of the generative trajectory.

\subsection{Generation Trajectory in Native Pixel Space}

Most pDiT models inherit their noise schedules, time parameterizations, and sampling strategies from conventional diffusion models or latent-space models. However, native pixel space has frequency structures and optimization properties that differ from those of compact latent spaces. Global contours, semantic layouts, and high-frequency textures do not form at the same rate along the diffusion trajectory. Thus, a unified noise process or linear transport path may not adequately fit all visual components.
Future research should revisit generation trajectories in pixel space from the perspectives of frequency, semantics, and spatial scale, enabling models to first establish stable global structures and then progressively recover high-frequency details. Sampling budgets should also shift from fixed allocation to sample-adaptive allocation, dynamically deciding whether to continue computation according to image clarity, semantic completeness, and local uncertainty. In this way, pixel generation may evolve from a fixed-step numerical integration process into an adaptive generation process driven by visual content and task requirements.

\subsection{Post-Training of pDiT}

Existing reinforcement learning (RL)~\cite{zhao2022alphaholdem,gan2024reflective,gan2024transductive} and preference optimization methods for diffusion models are primarily built on latent-space generative models~\cite{black2024training,wallace2024diffusion,yanintdiff}. A mature post-training paradigm specifically designed for native pDiT has not yet emerged, and related research still lacks systematic synthesis. Because pDiT's generative objective is defined directly in the final pixel space, reward signals can more directly evaluate fine-grained attributes including text rendering, structural preservation, local editing, and texture restoration, thereby reducing the supervision gap between latent representations and final-image quality.

However, high-resolution pixel trajectories substantially increase the cost of online sampling, reward evaluation, and policy updates. Existing VLM-based reward models are generally better at assessing high-level semantics and overall aesthetics, but may be insufficiently sensitive to local structures and high-frequency visual errors. They may even induce the model to obtain spuriously high rewards through shortcuts in sharpness, color, or texture. Excessive preference optimization may also compress pixel-level variation and reduce generative diversity.

Therefore, post-training for pDiT cannot simply copy the training recipes used for latent diffusion. Future work should focus on low-cost rollouts, dense feedback for local regions, reward models that account for both semantics and high-frequency details, and optimization mechanisms that suppress reward hacking and diversity degradation. How to coordinate exploration and exploitation, so that the model improves target attributes while preserving the original generative distribution, will be an important issue for post-training in pixel space.

\subsection{Static Distillation toward Non-Intrusive Representation }

A frozen vision foundation model can provide pDiT with priors such as object semantics, spatial relationships, and structural boundaries. However, discriminative models and generative models do not require the same visual representations. Vision foundation models usually emphasize semantic invariance, whereas image generation must also preserve color, texture, and local randomness. Direct token-wise matching of teacher features may compress the representational freedom of the generative model and cause discriminative semantic supervision to interfere excessively with detail restoration.

Future representation guidance should move from static, rigid feature matching toward non-intrusive and calibratable relational constraints. Teacher models are better suited to providing spatial layouts, regional relationships, and semantic boundaries, rather than prescribing the channel representations of the generative backbone. The strength of guidance should also vary dynamically with the diffusion stage and visual scale: it should provide necessary support during semantic-structure formation and gradually weaken the constraint during detail restoration. How to automatically determine when to guide, which regions to guide, and how strong the constraint should be is the central problem to be solved in this direction.

\subsection{Redefining the Computation–Fidelity Pareto Frontier}

The practical value of pDiT cannot be judged solely by a single FID score, CLIPScore, or inference latency. Although pixel-space models eliminate the fixed VAE, they must handle longer visual sequences and higher backbone computation costs. Fair comparison should therefore cover the full training and deployment lifecycle, including tokenizer pretraining cost, backbone training cost, peak memory usage, sampling cost, and quality across different dimensions of visual capability.

Future evaluation should place greater emphasis on capabilities that traditional metrics cannot fully capture, such as text readability, spatial layout, background consistency, texture fidelity, local editing, and reference-image preservation. Only when pDiT, under unified data, model scale, training time, and sampling budgets, establishes advantages in one or more key quality dimensions that cannot be dominated by latent-space models can its additional computational investment be considered fully justified. Establishing standardized computation–fidelity benchmarks across different resolutions and application scenarios is an important prerequisite for assessing the practical value of the pixel-space route.

\subsection{Shared Tokens toward Unified Understanding--Generation}

A shared token space does not mean understanding and generation have already been unified. Text understanding relies on abstract semantics, causal relations, and discrete symbolic reasoning, whereas pixel generation focuses more on spatial locality, continuous noise modeling, and high-frequency detail restoration. If all modalities are forced to share exactly the same representation granularity and computation path, gradient competition across tasks may arise and may damage either the original language capability or image generation quality.
Future unified models need to balance a shared backbone with modality-specific differences. Text, generated images, reference images, and task conditions can enter a unified context, while different modalities should still be allowed to use attention patterns, input projections, and output objectives suited to their own properties. The training process must also dynamically coordinate understanding, generation, editing, and reasoning tasks to avoid forgetting existing functions when new capabilities are introduced. True unification does not mean eliminating all structural differences; rather, it means preserving the necessary modeling mechanisms for different modalities on the basis of shared semantics and reasoning.

\subsection{Data and Evaluation Frameworks for Fine-Grained Visual}

The potential advantages of pixel-space methods mainly appear in tasks that traditional generation metrics cannot sufficiently measure, such as complex text rendering, precise local editing, identity preservation, reference-image control, and multi-image contextual reasoning. Existing datasets usually focus more on overall image quality and high-level semantic consistency, while providing insufficient coverage of fine edges, regular structures, repetitive textures, and preservation of non-edited regions.
Future data and evaluation frameworks should examine both semantic understanding and pixel-level fidelity. Test samples should cover scenarios such as long text and complex typography, fine-grained structures and regular patterns, local editing with background preservation, reference-subject consistency, and multi-image relational reasoning. Corresponding metrics should also extend beyond a single perceptual-quality measure to include OCR accuracy, structural preservation rate, edit locality, reference consistency, reasoning accuracy, and generation quality under a unit computational budget. Only by establishing systematic evaluation for fine-grained visual evidence can we accurately determine whether pDiT truly breaks through the capability boundaries imposed by latent-space compression.

\subsection{Summary}

Future research on pixel-space diffusion should not be understood as a return to early high-cost pixel denoising. Instead, it should be understood as a redesign of the representation interfaces, computation paths, and joint training mechanisms of visual foundation models. The most promising direction is not to perform full global attention over all pixels at every timestep, but to use dynamic token granularity, multi-scale continuous trajectories, non-intrusive semantic guidance, and task-aware computation allocation under end-to-end pixel supervision, so that the model preserves pixel-level details when needed and forms high-level semantic abstractions when needed.
In this sense, the long-term goal of pDiT is not simply to replace latent diffusion, but to build a visual foundation model that can simultaneously support high-fidelity generation, fine-grained control, cross-modal understanding, and contextual reasoning. Whether this goal can be achieved depends not only on larger models and more computational resources, but also on whether scalable structured computation mechanisms can be established in high-dimensional pixel space, and whether generation quality and understanding capability can be stably coordinated during unified training.

{
\small
\bibliographystyle{IEEEtran}
\bibliography{7-reference}

@String(AAAI = {AAAI})

@inproceedings{yan2025entropy,
  title={Entropy-Adaptive Diffusion Policy Optimization with Dynamic Step Alignment},
  author={Yan, RenYe and Cheng, Jikang and Gan, Yaozhong and Sun, Shikun and Wu, You and Yang, Yunfan and Ling, Liang and Lin, Jinlong and Zhu, Yeshuang and Zhou, Jie and others},
  booktitle={Proceedings of the IEEE/CVF International Conference on Computer Vision},
  pages={1924--1934},
  year={2025}
}

@inproceedings{yan2026less,
  title={Do Less, Achieve More: Do We Need Every-Step Optimization for RL Fine-tuning of Diffusion Models?},
  author={Yan, Renye and Cheng, Jikang and Sun, Shikun and Sun, Yi and Wu, You and Peng, Wei and Wang, Zongwei and Liang, Ling and Xing, Junliang and Cai, Yimao},
  booktitle={Proceedings of the IEEE/CVF Conference on Computer Vision and Pattern Recognition},
  pages={16561--16571},
  year={2026}
}

@inproceedings{rombach2022high,
  title={High-resolution image synthesis with latent diffusion models},
  author={Rombach, Robin and Blattmann, Andreas and Lorenz, Dominik and Esser, Patrick and Ommer, Bj{\"o}rn},
  booktitle={Proceedings of the IEEE/CVF conference on computer vision and pattern recognition},
  pages={10684--10695},
  year={2022}
}

@inproceedings{podell2024sdxl,
  title={Sdxl: Improving latent diffusion models for high-resolution image synthesis},
  author={Podell, Dustin and English, Zion and Lacey, Kyle and Blattmann, Andreas and Dockhorn, Tim and M{\"u}ller, Jonas and Penna, Joe and Rombach, Robin},
  booktitle={International Conference on Learning Representations},
  volume={2024},
  pages={1862--1874},
  year={2024}
}

@inproceedings{blattmann2023align,
  title={Align your latents: High-resolution video synthesis with latent diffusion models},
  author={Blattmann, Andreas and Rombach, Robin and Ling, Huan and Dockhorn, Tim and Kim, Seung Wook and Fidler, Sanja and Kreis, Karsten},
  booktitle={Proceedings of the IEEE/CVF conference on computer vision and pattern recognition},
  pages={22563--22575},
  year={2023}
}

@article{blattmann2023stable,
  title={Stable video diffusion: Scaling latent video diffusion models to large datasets},
  author={Blattmann, Andreas and Dockhorn, Tim and Kulal, Sumith and Mendelevitch, Daniel and Kilian, Maciej and Lorenz, Dominik and Levi, Yam and English, Zion and Voleti, Vikram and Letts, Adam and others},
  journal={arXiv preprint arXiv:2311.15127},
  year={2023}
}

@article{kwon2022diffusion,
  title={Diffusion models already have a semantic latent space},
  author={Kwon, Mingi and Jeong, Jaeseok and Uh, Youngjung},
  journal={arXiv preprint arXiv:2210.10960},
  year={2022}
}

@inproceedings{corneanu2024latentpaint,
  title={Latentpaint: Image inpainting in latent space with diffusion models},
  author={Corneanu, Ciprian and Gadde, Raghudeep and Martinez, Aleix M},
  booktitle={Proceedings of the IEEE/CVF winter conference on applications of computer vision},
  pages={4334--4343},
  year={2024}
}

@article{dar2026unconditional,
  title={Unconditional latent diffusion models memorize patient imaging data},
  author={Dar, Salman Ul Hassan and Seyfarth, Marvin and Ayx, Isabelle and Papavassiliu, Theano and Schoenberg, Stefan O and Siepmann, Robert Malte and Laqua, Fabian Christopher and Kahmann, Jannik and Frey, Norbert and Bae{\ss}ler, Bettina and others},
  journal={Nature biomedical engineering},
  volume={10},
  number={3},
  pages={458--472},
  year={2026},
  publisher={Nature Publishing Group UK London}
}

@article{heitmann2026picture,
  title={Picture perfect: Engaging customers with visual generative ai},
  author={Heitmann, Mark and Jansen, Tijmen PJ and Reisenbichler, Martin and Schweidel, David A},
  journal={Journal of Marketing},
  volume={90},
  number={4},
  pages={74--96},
  year={2026},
  publisher={SAGE Publications Sage CA: Los Angeles, CA}
}

@inproceedings{dufour2025around,
  title={Around the world in 80 timesteps: A generative approach to global visual geolocation},
  author={Dufour, Nicolas and Kalogeiton, Vicky and Picard, David and Landrieu, Loic},
  booktitle={Proceedings of the Computer Vision and Pattern Recognition Conference},
  pages={23016--23026},
  year={2025}
}

@article{yan2024exploration,
  title={The Exploration-Exploitation Dilemma Revisited: An Entropy Perspective},
  author={Yan, Renye and Gan, Yaozhong and Wu, You and Liang, Ling and Xing, Junliang and Cai, Yimao and Huang, Ru},
  journal={arXiv preprint arXiv:2408.09974},
  year={2024}
}

@article{sonderby2016ladder,
  title={Ladder variational autoencoders},
  author={S{\o}nderby, Casper Kaae and Raiko, Tapani and Maal{\o}e, Lars and S{\o}nderby, S{\o}ren Kaae and Winther, Ole},
  journal={Advances in neural information processing systems},
  volume={29},
  year={2016}
}

@inproceedings{ramchandran2021longitudinal,
  title={Longitudinal variational autoencoder},
  author={Ramchandran, Siddharth and Tikhonov, Gleb and Kujanp{\"a}{\"a}, Kalle and Koskinen, Miika and L{\"a}hdesm{\"a}ki, Harri},
  booktitle={International Conference on Artificial Intelligence and Statistics},
  pages={3898--3906},
  year={2021},
  organization={PMLR}
}

@inproceedings{kusner2017grammar,
  title={Grammar variational autoencoder},
  author={Kusner, Matt J and Paige, Brooks and Hern{\'a}ndez-Lobato, Jos{\'e} Miguel},
  booktitle={International conference on machine learning},
  pages={1945--1954},
  year={2017},
  organization={PMLR}
}

@inproceedings{khattar2019mvae,
  title={Mvae: Multimodal variational autoencoder for fake news detection},
  author={Khattar, Dhruv and Goud, Jaipal Singh and Gupta, Manish and Varma, Vasudeva},
  booktitle={The world wide web conference},
  pages={2915--2921},
  year={2019}
}

@article{vafaii2024poisson,
  title={Poisson variational autoencoder},
  author={Vafaii, Hadi and Galor, Dekel and Yates, Jacob L},
  journal={Advances in Neural Information Processing Systems},
  volume={37},
  pages={44871--44906},
  year={2024}
}

@article{abdulganiyu2025xidintfl,
  title={XIDINTFL-VAE: XGBoost-based intrusion detection of imbalance network traffic via class-wise focal loss variational autoencoder},
  author={Abdulganiyu, Oluwadamilare Harazeem and Tchakoucht, Taha Ait and Saheed, Yakub Kayode and Ahmed, Hilali Alaoui},
  journal={The Journal of Supercomputing},
  volume={81},
  number={1},
  pages={16},
  year={2025},
  publisher={Springer}
}

@article{zhang2026remaining,
  title={Remaining useful life prediction based on interpretable serialized variational autoencoder: A drift-diffusion stochastic equation perspective},
  author={Zhang, Jiusi and Chen, Kai and He, Renjun and Huang, Tenglong and Tian, Jilun and Wu, Shimeng and Yan, Pengfei and Cheng, Yuhua},
  journal={IEEE Transactions on Industrial Informatics},
  year={2026},
  publisher={IEEE}
}

@article{bonnaire2026diffusion,
  title={Why diffusion models don’t memorize: The role of implicit dynamical regularization in training},
  author={Bonnaire, Tony and Urfin, Rapha{\"e}l and Biroli, Giulio and M{\'e}zard, Marc},
  journal={Advances in Neural Information Processing Systems},
  volume={38},
  pages={141266--141286},
  year={2026}
}

@article{kim2026klass,
  title={Klass: Kl-guided fast inference in masked diffusion models},
  author={Kim, Seo Hyun and Hong, Sunwoo and Jung, Hojung and Park, Youngrok and Yun, Se-Young},
  journal={Advances in Neural Information Processing Systems},
  volume={38},
  pages={92267--92301},
  year={2026}
}

@article{zampini2026training,
  title={Training-free constrained generation with stable diffusion models},
  author={Zampini, Stefano and Christopher, Jacob K and Oneto, Luca and Anguita, Davide and Fioretto, Ferdinando},
  journal={Advances in Neural Information Processing Systems},
  volume={38},
  pages={27285--27316},
  year={2026}
}

@inproceedings{peebles2023scalable,
  title={Scalable diffusion models with transformers},
  author={Peebles, William and Xie, Saining},
  booktitle={Proceedings of the IEEE/CVF international conference on computer vision},
  pages={4195--4205},
  year={2023}
}

@inproceedings{ma2024sit,
  title={Sit: Exploring flow and diffusion-based generative models with scalable interpolant transformers},
  author={Ma, Nanye and Goldstein, Mark and Albergo, Michael S and Boffi, Nicholas M and Vanden-Eijnden, Eric and Xie, Saining},
  booktitle={European Conference on Computer Vision},
  pages={23--40},
  year={2024},
  organization={Springer}
}

@inproceedings{crowson2024scalable,
  title={Scalable high-resolution pixel-space image synthesis with hourglass diffusion transformers},
  author={Crowson, Katherine and Baumann, Stefan Andreas and Birch, Alex and Abraham, Tanishq Mathew and Kaplan, Daniel Z and Shippole, Enrico},
  booktitle={Forty-first International Conference on Machine Learning},
  year={2024}
}

@article{atzmon2024edify,
  title={Edify image: High-quality image generation with pixel space laplacian diffusion models},
  author={Atzmon, Yuval and Bala, Maciej and Balaji, Yogesh and Cai, Tiffany and Cui, Yin and Fan, Jiaojiao and Ge, Yunhao and Gururani, Siddharth and Huffman, Jacob and Isaac, Ronald and others},
  journal={arXiv preprint arXiv:2411.07126},
  year={2024}
}

@inproceedings{ma2026deco,
  title={Deco: Frequency-decoupled pixel diffusion for end-to-end image generation},
  author={Ma, Zehong and Wei, Longhui and Wang, Shuai and Zhang, Shiliang and Tian, Qi},
  booktitle={Proceedings of the IEEE/CVF Conference on Computer Vision and Pattern Recognition},
  pages={43600--43610},
  year={2026}
}

@article{ma2026pixelgen,
  title={PixelGen: Pixel Diffusion Beats Latent Diffusion with Perceptual Loss},
  author={Ma, Zehong and Xu, Ruihan and Zhang, Shiliang},
  journal={arXiv preprint arXiv:2602.02493},
  year={2026}
}

@article{wang2025pixnerd,
  title={Pixnerd: Pixel neural field diffusion},
  author={Wang, Shuai and Gao, Ziteng and Zhu, Chenhui and Huang, Weilin and Wang, Limin},
  journal={arXiv preprint arXiv:2507.23268},
  year={2025}
}

@inproceedings{yu2026pixeldit,
  title={Pixeldit: Pixel diffusion transformers for image generation},
  author={Yu, Yongsheng and Xiong, Wei and Nie, Weili and Sheng, Yichen and Liu, Shiqiu and Luo, Jiebo},
  booktitle={Proceedings of the IEEE/CVF Conference on Computer Vision and Pattern Recognition},
  pages={14273--14282},
  year={2026}
}

@article{baade2026latent,
  title={Latent forcing: Reordering the diffusion trajectory for pixel-space image generation},
  author={Baade, Alan and Chan, Eric Ryan and Sargent, Kyle and Chen, Changan and Johnson, Justin and Adeli, Ehsan and Fei-Fei, Li},
  journal={arXiv preprint arXiv:2602.11401},
  year={2026}
}

@inproceedings{bradbury2026your,
  title={Your latent mask is wrong: Pixel-equivalent latent compositing for diffusion models},
  author={Bradbury, Rowan and Zhong, Dazhi},
  booktitle={Proceedings of the IEEE/CVF Conference on Computer Vision and Pattern Recognition},
  pages={18630--18639},
  year={2026}
}

@inproceedings{mukhopadhyay2026scale,
  title={Scale space diffusion},
  author={Mukhopadhyay, Soumik and Udhayanan, Prateksha and Shrivastava, Abhinav},
  booktitle={Proceedings of the IEEE/CVF Conference on Computer Vision and Pattern Recognition},
  pages={35851--35860},
  year={2026}
}

@inproceedings{nguyen2026pixel,
  title={Pixel motion diffusion is what we need for robot control},
  author={Nguyen, E-Ro and Zhang, Yichi and Ranasinghe, Kanchana and Li, Xiang and Ryoo, Michael S},
  booktitle={Proceedings of the IEEE/CVF Conference on Computer Vision and Pattern Recognition},
  pages={23663--23672},
  year={2026}
}

@inproceedings{hoogeboom2025simpler,
  title={Simpler Diffusion: 1.5 FID on ImageNet512 with pixel-space diffusion},
  author={Hoogeboom, Emiel and Mensink, Thomas and Heek, Jonathan and Lamerigts, Kay and Gao, Ruiqi and Salimans, Tim},
  booktitle={Proceedings of the Computer Vision and Pattern Recognition Conference},
  pages={18062--18071},
  year={2025}
}

@article{zhang2024pixel,
  title={Pixel-space post-training of latent diffusion models},
  author={Zhang, Christina and Motwani, Simran and Yu, Matthew and Hou, Ji and Juefei-Xu, Felix and Tsai, Sam and Vajda, Peter and He, Zijian and Wang, Jialiang},
  journal={arXiv preprint arXiv:2409.17565},
  year={2024}
}

@inproceedings{li2026back,
  title={Back to basics: Let denoising generative models denoise},
  author={Li, Tianhong and He, Kaiming},
  booktitle={Proceedings of the IEEE/CVF Conference on Computer Vision and Pattern Recognition},
  pages={36115--36125},
  year={2026}
}

@article{ho2020denoising,
  title={Denoising diffusion probabilistic models},
  author={Ho, Jonathan and Jain, Ajay and Abbeel, Pieter},
  journal={Advances in neural information processing systems},
  volume={33},
  pages={6840--6851},
  year={2020}
}

@article{cai2026hidream,
  title={Hidream-o1-image: A natively unified image generative foundation model with pixel-level unified transformer},
  author={Cai, Qi and Chen, Jingwen and Gao, Chengmin and Gong, Zijian and Li, Yehao and Pan, Yingwei and Peng, Yi and Qiu, Zhaofan and Yu, Kai and Zhang, Yiheng and others},
  journal={arXiv preprint arXiv:2605.11061},
  year={2026}
}

@inproceedings{xie2025show,
  title={Show-o: One single transformer to unify multimodal understanding and generation},
  author={Xie, Jinheng and Mao, Weijia and Bai, Zechen and Zhang, David Junhao and Wang, Weihao and Lin, Kevin Qinghong and Gu, Yuchao and Chen, Zhijie and Yang, Zhenheng and Shou, Mike Zheng},
  booktitle={International Conference on Learning Representations},
  volume={2025},
  pages={28240--28264},
  year={2025}
}

@inproceedings{zhou2025transfusion,
  title={Transfusion: Predict the next token and diffuse images with one multi-modal model},
  author={Zhou, Chunting and Yu, Lili and Babu, Arun and Tirumala, Kushal and Yasunaga, Michihiro and Shamis, Leonid and Kahn, Jacob and Ma, Xuezhe and Zettlemoyer, Luke and Levy, Omer},
  booktitle={International Conference on Learning Representations},
  volume={2025},
  pages={6446--6469},
  year={2025}
}

@inproceedings{radford2021learning,
  title={Learning transferable visual models from natural language supervision},
  author={Radford, Alec and Kim, Jong Wook and Hallacy, Chris and Ramesh, Aditya and Goh, Gabriel and Agarwal, Sandhini and Sastry, Girish and Askell, Amanda and Mishkin, Pamela and Clark, Jack and others},
  booktitle={International conference on machine learning},
  pages={8748--8763},
  year={2021},
  organization={PmLR}
}

@article{oquab2023dinov2,
  title={Dinov2: Learning robust visual features without supervision},
  author={Oquab, Maxime and Darcet, Timoth{\'e}e and Moutakanni, Th{\'e}o and Vo, Huy and Szafraniec, Marc and Khalidov, Vasil and Fernandez, Pierre and Haziza, Daniel and Massa, Francisco and El-Nouby, Alaaeldin and others},
  journal={arXiv preprint arXiv:2304.07193},
  year={2023}
}

@inproceedings{hoogeboom2023simple,
  title={simple diffusion: End-to-end diffusion for high resolution images},
  author={Hoogeboom, Emiel and Heek, Jonathan and Salimans, Tim},
  booktitle={International Conference on Machine Learning},
  pages={13213--13232},
  year={2023},
  organization={PMLR}
}

@inproceedings{cheng2026sanity,
  title={A Sanity Check for Multi-In-Domain Face Forgery Detection in the Real World},
  author={Cheng, Jikang and Yan, Renye and Yan, Zhiyuan and Gan, Yaozhong and Zhang, Xueyi and Wang, Zhongyuan and Peng, Wei and Liang, Ling},
  booktitle={Proceedings of the IEEE/CVF Conference on Computer Vision and Pattern Recognition},
  pages={21306--21315},
  year={2026}
}

@article{xu2026edge,
  title={Edge deep learning in computer vision and medical diagnostics: a comprehensive survey},
  author={Xu, Yiwen and Khan, Tariq M and Song, Yang and Meijering, Erik},
  journal={arXiv preprint arXiv:2605.06714},
  year={2026}
}

@article{kage2026review,
  title={A review of pseudo-labeling for computer vision},
  author={Kage, Patrick and Rothenberger, Jay and Andreadis, Pavlos and Diochnos, Dimitrios},
  journal={Journal of Artificial Intelligence Research},
  volume={85},
  year={2026}
}

@inproceedings{ballesteros2026intelligent,
  title={Intelligent Recognition of Emergency Vehicles in Congested Traffic using Computer Vision},
  author={Ballesteros, Kyle Philip M and Cruz, Charles Matthew L Dela and Magbanua, Jewl Danielle R and Mancenido, Miguel Jose M and Comia, Lysa V},
  booktitle={2026 6th International Conference on Image Processing and Capsule Networks (ICIPCN)},
  pages={34--40},
  year={2026},
  organization={IEEE}
}

@article{ji2026computer,
  title={in Computer Vision},
  author={Ji, Yu and Wu, Wenzhi and Chen, Hang and Liu, Zhengjun},
  journal={Artificial Intelligence in Digital Image Processing: Theories, Methods, and Applications},
  pages={79},
  year={2026},
  publisher={Springer Nature}
}

@article{guo2022attention,
  title={Attention mechanisms in computer vision: A survey},
  author={Guo, Meng-Hao and Xu, Tian-Xing and Liu, Jiang-Jiang and Liu, Zheng-Ning and Jiang, Peng-Tao and Mu, Tai-Jiang and Zhang, Song-Hai and Martin, Ralph R and Cheng, Ming-Ming and Hu, Shi-Min},
  journal={Computational visual media},
  volume={8},
  number={3},
  pages={331--368},
  year={2022},
  publisher={TUP}
}

@article{wang2021generative,
  title={Generative adversarial networks in computer vision: A survey and taxonomy},
  author={Wang, Zhengwei and She, Qi and Ward, Tomas E},
  journal={ACM Computing Surveys (CSUR)},
  volume={54},
  number={2},
  pages={1--38},
  year={2021},
  publisher={ACM New York, NY, USA}
}

@article{bond2021deep,
  title={Deep generative modelling: A comparative review of vaes, gans, normalizing flows, energy-based and autoregressive models},
  author={Bond-Taylor, Sam and Leach, Adam and Long, Yang and Willcocks, Chris G},
  journal={IEEE transactions on pattern analysis and machine intelligence},
  volume={44},
  number={11},
  pages={7327--7347},
  year={2021},
  publisher={IEEE}
}

@article{zhao2016energy,
  title={Energy-based generative adversarial network},
  author={Zhao, Junbo and Mathieu, Michael and LeCun, Yann},
  journal={arXiv preprint arXiv:1609.03126},
  year={2016}
}

@article{dinh2014nice,
  title={Nice: Non-linear independent components estimation},
  author={Dinh, Laurent and Krueger, David and Bengio, Yoshua},
  journal={arXiv preprint arXiv:1410.8516},
  year={2014}
}

@article{van2016wavenet,
  title={Wavenet: A generative model for raw audio},
  author={Van Den Oord, Aaron and Dieleman, Sander and Zen, Heiga and Simonyan, Karen and Vinyals, Oriol and Graves, Alex and Kalchbrenner, Nal and Senior, Andrew and Kavukcuoglu, Koray and others},
  journal={arXiv preprint arXiv:1609.03499},
  volume={12},
  number={1},
  year={2016}
}

@inproceedings{zhu2017unpaired,
  title={Unpaired image-to-image translation using cycle-consistent adversarial networks},
  author={Zhu, Jun-Yan and Park, Taesung and Isola, Phillip and Efros, Alexei A},
  booktitle={Proceedings of the IEEE international conference on computer vision},
  pages={2223--2232},
  year={2017}
}

@article{croitoru2023diffusion,
  title={Diffusion models in vision: A survey},
  author={Croitoru, Florinel-Alin and Hondru, Vlad and Ionescu, Radu Tudor and Shah, Mubarak},
  journal={IEEE transactions on pattern analysis and machine intelligence},
  volume={45},
  number={9},
  pages={10850--10869},
  year={2023},
  publisher={Ieee}
}

@article{zou2026mixture,
  title={Mixture of global and local experts with diffusion transformer for controllable face generation},
  author={Zou, Xuechao and Zhang, Shun and Fu, Xing and Li, Yue and Li, Kai and Cao, Yushe and Lang, Congyan and Tao, Pin and Xing, Junliang},
  journal={IEEE Transactions on Pattern Analysis and Machine Intelligence},
  year={2026},
  publisher={IEEE}
}

@article{wang2026remasking,
  title={Remasking discrete diffusion models with inference-time scaling},
  author={Wang, Guanghan and Schiff, Yair and Sahoo, Subham and Kuleshov, Volodymyr},
  journal={Advances in Neural Information Processing Systems},
  volume={38},
  pages={147282--147339},
  year={2026}
}

@inproceedings{ciubotariu2026low,
  title={Low Light Image Enhancement Challenge at NTIRE 2026},
  author={Ciubotariu, George and Rehman, Abdur and Dharejo, Fayaz Ali and Naqvi, Rizwan Ali and Conde, Marcos V and Timofte, Radu and Jin, Zhi and Wu, Hongjun and Zhang, Wenjian and Ye, Chang and others},
  booktitle={Proceedings of the IEEE/CVF Conference on Computer Vision and Pattern Recognition},
  pages={1741--1753},
  year={2026}
}

@article{hong2026hyperspectral,
  title={Hyperspectral imaging},
  author={Hong, Danfeng and Li, Chenyu and Yokoya, Naoto and Zhang, Bing and Jia, Xiuping and Plaza, Antonio and Gamba, Paolo and Benediktsson, Jon Atli and Chanussot, Jocelyn},
  journal={Nature Reviews Methods Primers},
  volume={6},
  number={1},
  pages={19},
  year={2026},
  publisher={Nature Publishing Group UK London}
}

@article{bolya2026perception,
  title={Perception encoder: The best visual embeddings are not at the output of the network},
  author={Bolya, Daniel and Huang, Po-Yao and Sun, Peize and Cho, Jang Hyun and Madotto, Andrea and Wei, Chen and Ma, Tengyu and Zhi, Jiale and Rajasegaran, Jathushan and Bangalath, Hanoona and others},
  journal={Advances in Neural Information Processing Systems},
  volume={38},
  pages={60884--60937},
  year={2026}
}

@inproceedings{moskalenko2026ntire,
  title={NTIRE 2026 Challenge on Video Saliency Prediction: Methods and Results},
  author={Moskalenko, Andrey and Bryncev, Alexey and Kosmynin, Ivan and Shilovskaya, Kira and Erofeev, Mikhail and Vatolin, Dmitry and Timofte, Radu and Wang, Kun and Hu, Yupeng and Li, Zhiran and others},
  booktitle={Proceedings of the IEEE/CVF Conference on Computer Vision and Pattern Recognition},
  pages={2193--2206},
  year={2026}
}

@inproceedings{gushchin2026ntire,
  title={NTIRE 2026 Challenge on Robust AI-Generated Image Detection in the Wild},
  author={Gushchin, Aleksandr and Abud, Khaled and Shumitskaya, Ekaterina and Filippov, Artem and Bychkov, Georgii and Lavrushkin, Sergey and Erofeev, Mikhail and Antsiferova, Anastasia and Chen, Changsheng and Tan, Shunquan and others},
  booktitle={Proceedings of the IEEE/CVF Conference on Computer Vision and Pattern Recognition},
  pages={1895--1913},
  year={2026}
}

@inproceedings{frans2025one,
  title={One step diffusion via shortcut models},
  author={Frans, Kevin and Hafner, Danijar and Levine, Sergey and Abbeel, Pieter},
  booktitle={International Conference on Learning Representations},
  volume={2025},
  pages={34668--34684},
  year={2025}
}

@article{kingma2013auto,
  title={Auto-encoding variational bayes},
  author={Kingma, Diederik P and Welling, Max},
  journal={arXiv preprint arXiv:1312.6114},
  year={2013}
}

@inproceedings{zou2026turbo,
  title={Turbo-vaed: Fast and stable transfer of video-vaes to mobile devices},
  author={Zou, Ya and Yao, Jingfeng and Yu, Siyuan and Zhang, Shuai and Liu, Wenyu and Wang, Xinggang},
  booktitle={Proceedings of the AAAI Conference on Artificial Intelligence},
  volume={40},
  number={16},
  pages={14086--14094},
  year={2026}
}

@article{li2026sparc3d,
  title={Sparc3d: Sparse representation and construction for high-resolution 3d shapes modeling},
  author={Li, Zhihao and Wang, Yufei and Zheng, Heliang and Luo, Yihao and Wen, Bihan},
  journal={Advances in Neural Information Processing Systems},
  volume={38},
  pages={118582--118600},
  year={2026}
}

@inproceedings{bi2026vision,
  title={Vision foundation models can be good tokenizers for latent diffusion models},
  author={Bi, Tianci and Zhang, Xiaoyi and Lu, Yan and Zheng, Nanning},
  booktitle={Proceedings of the IEEE/CVF Conference on Computer Vision and Pattern Recognition},
  pages={43310--43319},
  year={2026}
}

@inproceedings{chen2025softvq,
  title={Softvq-vae: Efficient 1-dimensional continuous tokenizer},
  author={Chen, Hao and Wang, Ze and Li, Xiang and Sun, Ximeng and Chen, Fangyi and Liu, Jiang and Wang, Jindong and Raj, Bhiksha and Liu, Zicheng and Barsoum, Emad},
  booktitle={Proceedings of the Computer Vision and Pattern Recognition Conference},
  pages={28358--28370},
  year={2025}
}

@article{jia2025mgvq,
  title={MGVQ: Could VQ-VAE beat VAE? a generalizable tokenizer with multi-group quantization},
  author={Jia, Mingkai and Yin, Wei and Hu, Xiaotao and Guo, Jiaxin and Guo, Xiaoyang and Zhang, Qian and Long, Xiao-Xiao and Tan, Ping},
  journal={arXiv preprint arXiv:2507.07997},
  year={2025}
}

@inproceedings{guo2026improving,
  title={Improving autoregressive image generation through coarse-to-fine token prediction},
  author={Guo, Ziyao and Zhang, Kaipeng and Shieh, Michael Qizhe},
  booktitle={Proceedings of the IEEE/CVF Conference on Computer Vision and Pattern Recognition},
  pages={1230--1239},
  year={2026}
}

@inproceedings{zhang2025tar3d,
  title={Tar3d: Creating high-quality 3d assets via next-part prediction},
  author={Zhang, Xuying and Liu, Yutong and Li, Yangguang and Zhang, Renrui and Liu, Yufei and Wang, Kai and Ouyang, Wanli and Xiong, Zhiwei and Gao, Peng and Hou, Qibin and others},
  booktitle={Proceedings of the IEEE/CVF International Conference on Computer Vision},
  pages={5134--5145},
  year={2025}
}

@article{chen2025pixelflow,
  title={Pixelflow: Pixel-space generative models with flow},
  author={Chen, Shoufa and Ge, Chongjian and Zhang, Shilong and Sun, Peize and Luo, Ping},
  journal={arXiv preprint arXiv:2504.07963},
  year={2025}
}

@inproceedings{chen2026dip,
  title={Dip: Taming diffusion models in pixel space},
  author={Chen, Zhennan and Zhu, Junwei and Chen, Xu and Zhang, Jiangning and Hu, Xiaobin and Zhao, Hanzhen and Wang, Chengjie and Yang, Jian and Tai, Ying},
  booktitle={Proceedings of the IEEE/CVF Conference on Computer Vision and Pattern Recognition},
  pages={36136--36146},
  year={2026}
}

@article{he2026hyperdit,
  title={Hyperdit: Hyper-connected transformers for high-fidelity pixel-space diffusion},
  author={He, Yu and Ma, Lichen and Guo, Zipeng and Shan, Xinyuan and Fu, Jingling and Chen, Dong and Huang, Junshi and Li, Yan},
  journal={arXiv preprint arXiv:2605.15741},
  year={2026}
}

@inproceedings{jain2023vectorfusion,
  title={Vectorfusion: Text-to-svg by abstracting pixel-based diffusion models},
  author={Jain, Ajay and Xie, Amber and Abbeel, Pieter},
  booktitle={Proceedings of the IEEE/CVF Conference on Computer Vision and Pattern Recognition},
  pages={1911--1920},
  year={2023}
}

@article{rozet2026lost,
  title={Lost in latent space: An empirical study of latent diffusion models for physics emulation},
  author={Rozet, Fran{\c{c}}ois and Ohana, Ruben and McCabe, Michael and Louppe, Gilles and Lanusse, Fran{\c{c}}ois and Ho, Shirley},
  journal={Advances in Neural Information Processing Systems},
  volume={38},
  pages={134612--134657},
  year={2026}
}

@inproceedings{cheng2025leanvae,
  title={Leanvae: An ultra-efficient reconstruction vae for video diffusion models},
  author={Cheng, Yu and Yuan, Fajie},
  booktitle={Proceedings of the IEEE/CVF International Conference on Computer Vision},
  pages={15692--15702},
  year={2025}
}

@article{lipman2022flow,
  title={Flow matching for generative modeling},
  author={Lipman, Yaron and Chen, Ricky TQ and Ben-Hamu, Heli and Nickel, Maximilian and Le, Matt},
  journal={arXiv preprint arXiv:2210.02747},
  year={2022}
}

@article{liu2022flow,
  title={Flow straight and fast: Learning to generate and transfer data with rectified flow},
  author={Liu, Xingchao and Gong, Chengyue and Liu, Qiang},
  journal={arXiv preprint arXiv:2209.03003},
  year={2022}
}

@article{song2019generative,
  title={Generative modeling by estimating gradients of the data distribution},
  author={Song, Yang and Ermon, Stefano},
  journal={Advances in neural information processing systems},
  volume={32},
  year={2019}
}

@article{song2020score,
  title={Score-based generative modeling through stochastic differential equations},
  author={Song, Yang and Sohl-Dickstein, Jascha and Kingma, Diederik P and Kumar, Abhishek and Ermon, Stefano and Poole, Ben},
  journal={arXiv preprint arXiv:2011.13456},
  year={2020}
}

@article{dhariwal2021diffusion,
  title={Diffusion models beat gans on image synthesis},
  author={Dhariwal, Prafulla and Nichol, Alexander},
  journal={Advances in neural information processing systems},
  volume={34},
  pages={8780--8794},
  year={2021}
}

@inproceedings{ronneberger2015u,
  title={U-net: Convolutional networks for biomedical image segmentation},
  author={Ronneberger, Olaf and Fischer, Philipp and Brox, Thomas},
  booktitle={International Conference on Medical image computing and computer-assisted intervention},
  pages={234--241},
  year={2015},
  organization={Springer}
}

@article{trebing2021smaat,
  title={SmaAt-UNet: Precipitation nowcasting using a small attention-UNet architecture},
  author={Trebing, Kevin and Staǹczyk, Tomasz and Mehrkanoon, Siamak},
  journal={Pattern Recognition Letters},
  volume={145},
  pages={178--186},
  year={2021},
  publisher={Elsevier}
}

@article{he2022swin,
  title={Swin transformer embedding UNet for remote sensing image semantic segmentation},
  author={He, Xin and Zhou, Yong and Zhao, Jiaqi and Zhang, Di and Yao, Rui and Xue, Yong},
  journal={IEEE transactions on geoscience and remote sensing},
  volume={60},
  pages={1--15},
  year={2022},
  publisher={IEEE}
}

@article{simeoni2025dinov3,
  title={Dinov3},
  author={Sim{\'e}oni, Oriane and Vo, Huy V and Seitzer, Maximilian and Baldassarre, Federico and Oquab, Maxime and Jose, Cijo and Khalidov, Vasil and Szafraniec, Marc and Yi, Seungeun and Ramamonjisoa, Micha{\"e}l and others},
  journal={arXiv preprint arXiv:2508.10104},
  year={2025}
}

@article{yu2024representation,
  title={Representation alignment for generation: Training diffusion transformers is easier than you think},
  author={Yu, Sihyun and Kwak, Sangkyung and Jang, Huiwon and Jeong, Jongheon and Huang, Jonathan and Shin, Jinwoo and Xie, Saining},
  journal={arXiv preprint arXiv:2410.06940},
  year={2024}
}

@article{shin2026representation,
  title={Representation alignment for just image transformers is not easier than you think},
  author={Shin, Jaeyo and Kim, Jiwook and Shim, Hyunjung},
  journal={arXiv preprint arXiv:2603.14366},
  year={2026}
}

@article{heusel2017gans,
  title={Gans trained by a two time-scale update rule converge to a local nash equilibrium},
  author={Heusel, Martin and Ramsauer, Hubert and Unterthiner, Thomas and Nessler, Bernhard and Hochreiter, Sepp},
  journal={Advances in neural information processing systems},
  volume={30},
  year={2017}
}

@article{zheng2025diffusion,
  title={Diffusion transformers with representation autoencoders},
  author={Zheng, Boyang and Ma, Nanye and Tong, Shengbang and Xie, Saining},
  journal={arXiv preprint arXiv:2510.11690},
  year={2025}
}

@inproceedings{yao2025reconstruction,
  title={Reconstruction vs. generation: Taming optimization dilemma in latent diffusion models},
  author={Yao, Jingfeng and Yang, Bin and Wang, Xinggang},
  booktitle={Proceedings of the Computer Vision and Pattern Recognition Conference},
  pages={15703--15712},
  year={2025}
}

@inproceedings{he2022masked,
  title={Masked autoencoders are scalable vision learners},
  author={He, Kaiming and Chen, Xinlei and Xie, Saining and Li, Yanghao and Doll{\'a}r, Piotr and Girshick, Ross},
  booktitle={Proceedings of the IEEE/CVF conference on computer vision and pattern recognition},
  pages={16000--16009},
  year={2022}
}

@inproceedings{zhai2023sigmoid,
  title={Sigmoid loss for language image pre-training},
  author={Zhai, Xiaohua and Mustafa, Basil and Kolesnikov, Alexander and Beyer, Lucas},
  booktitle={Proceedings of the IEEE/CVF international conference on computer vision},
  pages={11975--11986},
  year={2023}
}

@article{vaswani2017attention,
  title={Attention is all you need},
  author={Vaswani, Ashish and Shazeer, Noam and Parmar, Niki and Uszkoreit, Jakob and Jones, Llion and Gomez, Aidan N and Kaiser, {\L}ukasz and Polosukhin, Illia},
  journal={Advances in neural information processing systems},
  volume={30},
  year={2017}
}

@article{sener2018multi,
  title={Multi-task learning as multi-objective optimization},
  author={Sener, Ozan and Koltun, Vladlen},
  journal={Advances in neural information processing systems},
  volume={31},
  year={2018}
}

@article{lin2019pareto,
  title={Pareto multi-task learning},
  author={Lin, Xi and Zhen, Hui-Ling and Li, Zhenhua and Zhang, Qing-Fu and Kwong, Sam},
  journal={Advances in neural information processing systems},
  volume={32},
  year={2019}
}

@article{kaplan2020scaling,
  title={Scaling laws for neural language models},
  author={Kaplan, Jared and McCandlish, Sam and Henighan, Tom and Brown, Tom B and Chess, Benjamin and Child, Rewon and Gray, Scott and Radford, Alec and Wu, Jeffrey and Amodei, Dario},
  journal={arXiv preprint arXiv:2001.08361},
  year={2020}
}

@article{hoffmann2022training,
  title={Training compute-optimal large language models},
  author={Hoffmann, Jordan and Borgeaud, Sebastian and Mensch, Arthur and Buchatskaya, Elena and Cai, Trevor and Rutherford, Eliza and Casas, DDL and Hendricks, Lisa Anne and Welbl, Johannes and Clark, Aidan and others},
  journal={arXiv preprint arXiv:2203.15556},
  volume={10},
  year={2022}
}

@inproceedings{tan2019efficientnet,
  title={Efficientnet: Rethinking model scaling for convolutional neural networks},
  author={Tan, Mingxing and Le, Quoc},
  booktitle={International conference on machine learning},
  pages={6105--6114},
  year={2019},
  organization={PMLR}
}

@article{zhang2026spatial,
  title={Spatial Gram Alignment for Ultra-High-Resolution Image Synthesis},
  author={Zhang, Jinjin and Guo, Xiefan and Huang, Di},
  journal={arXiv preprint arXiv:2605.20808},
  year={2026}
}

@inproceedings{xiao2025omnigen,
  title={Omnigen: Unified image generation},
  author={Xiao, Shitao and Wang, Yueze and Zhou, Junjie and Yuan, Huaying and Xing, Xingrun and Yan, Ruiran and Li, Chaofan and Wang, Shuting and Huang, Tiejun and Liu, Zheng},
  booktitle={Proceedings of the IEEE/CVF Conference on Computer Vision and Pattern Recognition},
  pages={13294--13304},
  year={2025}
}

@article{wei2024general,
  title={General ocr theory: Towards ocr-2.0 via a unified end-to-end model},
  author={Wei, Haoran and Liu, Chenglong and Chen, Jinyue and Wang, Jia and Kong, Lingyu and Xu, Yanming and Ge, Zheng and Zhao, Liang and Sun, Jianjian and Peng, Yuang and others},
  journal={arXiv preprint arXiv:2409.01704},
  year={2024}
}

@article{wang2024instantid,
  title={Instantid: Zero-shot identity-preserving generation in seconds},
  author={Wang, Qixun and Bai, Xu and Wang, Haofan and Qin, Zekui and Chen, Anthony and Li, Huaxia and Tang, Xu and Hu, Yao},
  journal={arXiv preprint arXiv:2401.07519},
  year={2024}
}

@article{jiang2024mantis,
  title={Mantis: Interleaved multi-image instruction tuning},
  author={Jiang, Dongfu and He, Xuan and Zeng, Huaye and Wei, Cong and Ku, Max and Liu, Qian and Chen, Wenhu},
  journal={arXiv preprint arXiv:2405.01483},
  year={2024}
}

@inproceedings{saharia2022palette,
  title={Palette: Image-to-image diffusion models},
  author={Saharia, Chitwan and Chan, William and Chang, Huiwen and Lee, Chris and Ho, Jonathan and Salimans, Tim and Fleet, David and Norouzi, Mohammad},
  booktitle={ACM SIGGRAPH 2022 conference proceedings},
  pages={1--10},
  year={2022}
}

@article{shen2025generative,
  title={When Generative Replay Meets Evolving Deepfakes: Domain-Aware Relative Weighting for Incremental Face Forgery Detection},
  author={Shen, Hao and Cheng, Jikang and Yan, Renye and Wang, Zhongyuan and Peng, Wei and Huang, Baojin},
  journal={arXiv preprint arXiv:2511.18436},
  year={2025}
}

@article{deng2026blazeedit,
  title={BlazeEdit: Generalist Image Editing on Mobile Devices with Image-to-Image Diffusion Models},
  author={Deng, Fei and Xu, Yanwu and Bao, Zhipeng and Zhang, Zhixing and Jia, Haolin and Raveendran, Karthik and Wei, Jianing},
  journal={arXiv preprint arXiv:2605.28067},
  year={2026}
}

@article{cao2026pixgs,
  title={PixGS: Pixel-Space Diffusion for Direct 3D Gaussian Splat Generation},
  author={Cao, Duy and Nguyen-Ha, Phong},
  journal={arXiv preprint arXiv:2607.01803},
  year={2026}
}

@article{gao2026pixworld,
  title={PixWorld: Unifying 3D Scene Generation and Reconstruction in Pixel Space},
  author={Gao, Sensen and Wang, Zhaoqing and Cao, Qihang and Yu, Dongdong and Wang, Changhu and Bian, Jia-Wang},
  journal={arXiv preprint arXiv:2607.05373},
  year={2026}
}

@article{xu2026pointdit,
  title={PointDiT: Pixel-Space Diffusion for Monocular Geometry Estimation},
  author={Xu, Haofei and Wu, Rundi and Henzler, Philipp and Kalischek, Nikolai and Oechsle, Michael and Manhardt, Fabian and Pollefeys, Marc and Geiger, Andreas and Tombari, Federico and Niemeyer, Michael},
  journal={arXiv preprint arXiv:2607.02515},
  year={2026}
}

@article{xu2024medsyn,
  title={MedSyn: text-guided anatomy-aware synthesis of high-fidelity 3-D CT images},
  author={Xu, Yanwu and Sun, Li and Peng, Wei and Jia, Shuyue and Morrison, Katelyn and Perer, Adam and Zandifar, Afrooz and Visweswaran, Shyam and Eslami, Motahhare and Batmanghelich, Kayhan},
  journal={IEEE Transactions on Medical Imaging},
  volume={43},
  number={10},
  pages={3648--3660},
  year={2024},
  publisher={IEEE}
}

@article{zhang2026pixel,
  title={Pixel-Level Residual Diffusion Transformer: Scalable 3D CT Volume Generation},
  author={Zhang, Zhenkai and Hiller, Markus and Ehinger, Krista A and Drummond, Tom},
  journal={arXiv preprint arXiv:2606.20112},
  year={2026}
}

@article{chen2026ppdm,
  title={PPDM: Pixel Puzzling Diffusion Model for Speed and Memory Efficient Volumetric Medical Image Translation},
  author={Chen, Tianqi and Hou, Jun and Zhou, Yinchi and Duncan, James S and Liu, Chi and Zhou, Bo},
  journal={arXiv preprint arXiv:2606.15323},
  year={2026}
}

@inproceedings{black2024training,
  title={Training diffusion models with reinforcement learning},
  author={Black, Kevin and Janner, Michael and Du, Yilun and Kostrikov, Ilya and Levine, Sergey},
  booktitle={International Conference on Learning Representations},
  volume={2024},
  pages={4965--4987},
  year={2024}
}

@inproceedings{wallace2024diffusion,
  title={Diffusion model alignment using direct preference optimization},
  author={Wallace, Bram and Dang, Meihua and Rafailov, Rafael and Zhou, Linqi and Lou, Aaron and Purushwalkam, Senthil and Ermon, Stefano and Xiong, Caiming and Joty, Shafiq and Naik, Nikhil},
  booktitle={Proceedings of the IEEE/CVF Conference on Computer Vision and Pattern Recognition},
  pages={8228--8238},
  year={2024}
}

@article{karras2024guiding,
  title={Guiding a diffusion model with a bad version of itself},
  author={Karras, Tero and Aittala, Miika and Kynk{\"a}{\"a}nniemi, Tuomas and Lehtinen, Jaakko and Aila, Timo and Laine, Samuli},
  journal={Advances in Neural Information Processing Systems},
  volume={37},
  pages={52996--53021},
  year={2024}
}

@article{tevet2022human,
  title={Human motion diffusion model},
  author={Tevet, Guy and Raab, Sigal and Gordon, Brian and Shafir, Yonatan and Cohen-Or, Daniel and Bermano, Amit H},
  journal={arXiv preprint arXiv:2209.14916},
  year={2022}
}

@article{fuest2026diffusion,
  title={Diffusion models and representation learning: A survey},
  author={Fuest, Michael and Ma, Pingchuan and Gui, Ming and Schusterbauer, Johannes and Hu, Vincent Tao and Ommer, Bj{\"o}rn},
  journal={IEEE Transactions on Pattern Analysis and Machine Intelligence},
  year={2026},
  publisher={IEEE}
}

@inproceedings{karnewar2023holodiffusion,
  title={Holodiffusion: Training a 3d diffusion model using 2d images},
  author={Karnewar, Animesh and Vedaldi, Andrea and Novotny, David and Mitra, Niloy J},
  booktitle={Proceedings of the IEEE/CVF conference on computer vision and pattern recognition},
  pages={18423--18433},
  year={2023}
}

@article{lovelace2023latent,
  title={Latent diffusion for language generation},
  author={Lovelace, Justin and Kishore, Varsha and Wan, Chao and Shekhtman, Eliot and Weinberger, Kilian Q},
  journal={Advances in Neural Information Processing Systems},
  volume={36},
  pages={56998--57025},
  year={2023}
}

@article{xu2026pixel,
  title={Pixel-perfect depth with semantics-prompted diffusion transformers},
  author={Xu, Gangwei and Lin, Haotong and Luo, Hongcheng and Wang, Xianqi and Yao, Jingfeng and Zhu, Lianghui and Pu, Yuechuan and Chi\_, Cheng and Sun, Haiyang and Wang, Bing and others},
  journal={Advances in Neural Information Processing Systems},
  volume={38},
  pages={174731--174755},
  year={2026}
}

@inproceedings{elata2025novel,
  title={Novel view synthesis with pixel-space diffusion models},
  author={Elata, Noam and Kawar, Bahjat and Ostrovsky-Berman, Yaron and Farber, Miriam and Sokolovsky, Ron},
  booktitle={Proceedings of the IEEE/CVF Conference on Computer Vision and Pattern Recognition},
  pages={26756--26766},
  year={2025}
}

@inproceedings{si2024freeu,
  title={Freeu: Free lunch in diffusion u-net},
  author={Si, Chenyang and Huang, Ziqi and Jiang, Yuming and Liu, Ziwei},
  booktitle={Proceedings of the IEEE/CVF conference on computer vision and pattern recognition},
  pages={4733--4743},
  year={2024}
}

@article{tian2026u,
  title={U-repa: Aligning diffusion u-nets to vits},
  author={Tian, Yuchuan and Chen, Hanting and Zheng, Mengyu and Liang, Yuchen and Xu, Chao and Wang, Yunhe},
  journal={Advances in Neural Information Processing Systems},
  volume={38},
  pages={11003--11024},
  year={2026}
}

@inproceedings{leng2025repa,
  title={Repa-e: Unlocking vae for end-to-end tuning of latent diffusion transformers},
  author={Leng, Xingjian and Singh, Jaskirat and Hou, Yunzhong and Xing, Zhenchang and Xie, Saining and Zheng, Liang},
  booktitle={Proceedings of the IEEE/CVF International Conference on Computer Vision},
  pages={18262--18272},
  year={2025}
}

@article{dao2022flashattention,
  title={Flashattention: Fast and memory-efficient exact attention with io-awareness},
  author={Dao, Tri and Fu, Dan and Ermon, Stefano and Rudra, Atri and R{\'e}, Christopher},
  journal={Advances in neural information processing systems},
  volume={35},
  pages={16344--16359},
  year={2022}
}

@article{li2026videococo,
  title={VideoCoCo: Code-as-CoT for Physically-Consistent Video Generation via an Agentic Dual-Engine System},
  author={Li, Haodong and Ren, Tianfei and Ma, Xiaoxiao and Qing, Chunmei and Fang, Zhen and He, Sipeng and Guo, Ziyu and Wu, Haoyu and Tian, Juanxi and Zou, Yihang and others},
  journal={arXiv preprint arXiv:2607.27380},
  year={2026}
}

@article{yanmultitune,
  title={MultiTune: Phase-Aware Multi-Objective Optimization for Diffusion Models},
  author={Yan, Renye and Cheng, Jikang and Sun, Shikun and Sun, Yi and Peng, Wei and Gan, Yaozhong and Wu, You and Liang, Ling and Xing, Junliang and Cai, Yimao}
}

@article{yanintdiff,
  title={IntDiff: Mitigating Reward Hacking via Intrinsic rewards for Diffusion Model Fine-Tuning},
  author={Yan, Renye and Cheng, Jikang and Gan, Yaozhong and Wu, You and Peng, Wei and Liang, Ling and Xing, Junliang and Zhang, Jinchao and Cai, Yimao}
}

@article{lu2026semantic,
  title={Semantic Granularity Navigation in Image Editing},
  author={Lu, Liangsi and Guo, Minzhe and Chen, Xuhang and Shi, Yang},
  journal={arXiv preprint arXiv:2605.21190},
  year={2026}
}

@inproceedings{zhao2022alphaholdem,
  title={AlphaHoldem: High-performance artificial intelligence for heads-up no-limit poker via end-to-end reinforcement learning},
  author={Zhao, Enmin and Yan, Renye and Li, Jinqiu and Li, Kai and Xing, Junliang},
  booktitle={Proceedings of the AAAI conference on artificial intelligence},
  volume={36},
  number={4},
  pages={4689--4697},
  year={2022}
}

@article{gan2024reflective,
  title={Reflective policy optimization},
  author={Gan, Yaozhong and Yan, Renye and Wu, Zhe and Xing, Junliang},
  journal={arXiv preprint arXiv:2406.03678},
  year={2024}
}

@article{gan2024transductive,
  title={Transductive off-policy proximal policy optimization},
  author={Gan, Yaozhong and Yan, Renye and Tan, Xiaoyang and Wu, Zhe and Xing, Junliang},
  journal={arXiv preprint arXiv:2406.03894},
  year={2024}
}
}

\end{CJK*}
\end{document}